\documentclass[preprint,12pt,authoryear]{elsarticle}

\usepackage{textcomp, gensymb}

\usepackage{amssymb}
\usepackage{adjustbox}
\usepackage{mathtools}
\usepackage{gensymb}
\usepackage{graphicx}
\usepackage{caption,subcaption}
\usepackage{pgffor}
\usepackage{amsmath}
\usepackage{tikz-cd}
\usepackage{latexsym}
\usepackage{longtable}
\usepackage{lscape}

\usepackage{tikz}
\usetikzlibrary{decorations.pathreplacing}
\usetikzlibrary{fadings}
\usepackage{pdflscape}
\usepackage{multirow}  
\usepackage{booktabs}       

\journal{Computational Brain \& Behavior}

\begin{document}

\begin{frontmatter}

    \title{Convolutional Neural Networks Trained to Identify Words Provide a Surprisingly Good Account of Visual Form Priming Effects}

    \author[inst1]{Dong Yin}
    \author[inst1]{Valerio Biscione}
    \author[inst1]{Jeffrey S. Bowers}

    \affiliation[inst1]{organization={Department of Psychology},
        addressline={University of Bristol},
        city={Bristol},
        postcode={BS8 1TL},
        country={United Kingdom}}

    \begin{abstract}

        A wide variety of orthographic coding schemes and models of visual word identification have been developed to account for masked priming data that provide a measure of orthographic similarity between letter strings. These models tend to include hand-coded orthographic representations with single unit coding for specific forms of knowledge (e.g., units coding for a letter in a given position). Here we assess how well a range of these coding schemes and models account for the pattern of form priming effects taken from the Form Priming Project and compare these findings to results observed with 11 standard deep neural network models (DNNs) developed in computer science. We find that deep convolutional networks (CNNs) perform as well or better than the coding schemes and word recognition models, whereas transformer networks did less well. The success of CNNs is remarkable as their architectures were not developed to support word recognition (they were designed to perform well on object recognition), they classify pixel images of words (rather than artificial encodings of letter strings), and their training was highly simplified (not respecting many key aspects of human experience). In addition to these form priming effects, we find that the DNNs can account for visual similarity effects on priming that are beyond all current psychological models of priming. The findings add to the recent work of \citep{hannagan-2021} and suggest that CNNs should be given more attention in psychology as models of human visual word recognition.
        

    \end{abstract}


\end{frontmatter}
\tnotetext[1]{prova prova}

\section{Introduction} \label{sec:intro}

Skilled visual word identification requires extensive experience with written words. It entails identifying the visual characteristics of letters (e.g., oriented lines), mapping these features onto letters, coding for letter order, and eventually selecting word candidates from one's vocabulary \citep{carreiras-2014}. The process of representing the identity and orders of letters in a letter string is referred to as orthographic coding and it constitutes a crucial component of word identification, with different models of word identification adopting different orthographic coding schemes.

Much of the empirical research directed at distinguishing different orthographic coding schemes and models of word identification more generally comes from priming studies that vary the similarity of the prime and target. Various priming procedures have been used, but the most common is the masked Lexical Decision Task (LDT) as introduced by \citet{forster-1987}. The procedure involves measuring how quickly people classify a target stimulus as a word or nonword when it is briefly preceded by a prime. The basic finding is that orthographically similar prime strings speed responses to the targets relative to unrelated prime strings (e.g., \citealt{schoonbaert-2004, burt-2017, bhide-2014}). The assumption is that the greater the priming, the greater the orthographic similarity between the prime and the target. A variety of different models of letter coding and models of word identification more broadly have been developed in an attempt to account for more variance in masked form priming experiments.

The primary objective of the current study is to investigate to what extent artificial Deep Neural Network (DNN) models developed in computer science with architectures designed to classify images of objects can account for masked form priming data, and in addition, compare their successes to some standard models of orthographic coding and word recognition. In all, we test 11 DDNs (different versions of convolutional and transformer networks), five orthographic coding schemes, and three models of visual recognition, as well as five control conditions, as detailed below.

\subsection{Orthographic coding schemes and models of word identification} \label{sec:coding schemes and models}
Any model of word recognition needs to include a series of basic processes. This includes encoding the letters and their order, a process of mapping the ordered letters onto lexical representations, and finally a manner of selecting one lexical entry from others. Different models adopt different accounts of these basic processing steps. Most relevant for present purposes, models in the psychological literature have taken three basic approaches to encoding letters and letter orders, namely, slot-based coding, context-based coding, and context-independent coding.

On slot-based coding schemes, separate slots for position-specific letter codes are assumed. For example, the word CAT might be coded by activating the three letter codes C\textsubscript{1}, A\textsubscript{2}, and T\textsubscript{3}, whereas the word ACT would be coded as A\textsubscript{1}, C\textsubscript{2}, and T\textsubscript{3} (where the subscript indexes letter position). Because letter codes are conjunctions of letter identities and letter position, the letter A\textsubscript{1} and A\textsubscript{2} are simply different letters (accordingly, CAT and ACT only share one letter, namely T\textsubscript{3}).

In context-based coding schemes, letters are coded in relation to other letters in the letter string. For example, in open-bigram coding schemes (e.g., \citealt{grainger-2004}), a letter string is coded in terms of all of the ordered letter pairs that it contains. For example CAT is coded by the letter pairs CA, AT, and CT, whereas ACT is coded as AC, AT, and CT. Various different versions of context-based coding schemes (and different versions of open-bigrams) have been proposed, and these again impact how transposing letters and other manipulations impact the orthographic similarity between letter strings.

Finally, in context-independent coding schemes, letter units are coded independently of position and context. That is, a node that codes the letter A is activated when the input stimulus contains an A, irrespective of its serial position or surrounding context (the same C, A, and T letter units are part of CAT and ACT). The ordering of letters is computed on-line (in order to distinguish between CAT and ACT), and this is achieved in various ways. For example, in the spatial coding scheme \citep{davis-2010A}, the precise time at which units are activated codes for letter order. Again, various versions of context-independent coding schemes have been proposed with consequences for the orthographic similarity of letter strings.

These different orthographic coding schemes form the front end of more complete models of visual word identification that include processes that select word representations from these orthographic encodings. For example, the Interactive Activation (IA) Model \citep{mcclelland-1981}, the overlap model \citep{gomez-2008}, and the Bayesian Reader \citep{norris-2006} all adopt different versions of slot coding; the open bigram \citep{grainger-2004-model} and Seriol \citep{whitney-2001} models use different context-based encoding schemes; the spatial coding \citep{davis-2010A} and SOLAR models \citep{davis-2001-solar} uses context independent encoding scheme. Note that the predictions of masked priming in these models are the product of both the encoding schemes and the additional processes that support lexical selection. In addition to these models, the Letters in Time and Retinotopic Space (LTRS; \citealt{adelman-2011}) model is agnostic to the encoding scheme and instead makes predictions based on the rate at which different features of the stimulus (that could take different forms) are extracted. We will consider how well various orthographic coding schemes as well as models of word identification account for masked priming results reported in the form priming project \citep{adelman-2014}.

\subsection{Deep Neural Network} \label{subsection dnn}
DNNs are a type of artificial neural network in which the input and output layers are separated by multiple (hidden) layers, forming a hierarchical structure. Two of the most common types of DNNs are convolutional neural networks (CNNs) and transformers.

CNNs are inspired by biological vision (\citealt{felleman-1991, krubitzer-1993, sereno-2015}).  The convolutions in CNNs refer to a set of feature detectors that repeat at different spatial locations to produce a series of feature maps (analogous to simple cells in V1, for example, where the same feature detector -- e.g., a vertical line detector -- repeats at multiple retinal locations). The convolutions are followed by a pooling operation in which corresponding features in nearby spatial locations are mapped together (analogous to complex cells that map together corresponding simple cells in nearby retinal locations), after which more feature detectors and pooling operations are applied, hierarchically, to form more and more complex feature detectors that are more invariant to spatial location. Different CNNs differ in various ways, including the number of hidden layers (many models have over 100 hidden layers), but they generally include ``localist'' or ``one hot'' representations in the output layer such that a single output unit codes for a specific category (e.g., an object class such as banana, or in the current case, a specific word), and CNNs tend to be trained through back-propagation as is the case with older parallel distributed processing models \citep{Rumelhart1986}.



In contrast, Vision Transformers (ViTs; \citealt{dosovitskiy-2020}) do not include any convolutions but instead introduce a self-attention mechanism. In object classification, this mechanism divides the input image into patches, and a similarity score between each and every patch is computed. These scores are used to compute a new representation of the image that emphasizes the most relevant features for the task at hand. Overall these models have much more complicated architectures, and describing them is beyond the scope of this paper. But again, the models include multiple layers, tend to include localist output codes, and are trained with back-propagation.


Importantly for present purposes, CNNs and ViTs are not only highly successful engineering tools that support a wide range of challenging AI tasks, but they are also often claimed to provide good models of the human visual system, and indeed, they are the most successful models in predicting judgements of category typicality \citep{lake-2015} and predicting object classification errors \citep{jozwik-2017} and human similarity judgements for natural images \citep{Peterson2018} on several datasets. DNNs have also been good at predicting the neural activation patterns elicited during object recognition in both human and non-human primates' ventral visual processing streams \citep{cichy-2016, storrs-2021}. A benchmark called Brain-Score has been developed to assess similarities between biological visual systems and DNNs \citep{schrimpf-2018}. The best performing models on the Brain-Score benchmarks are often described as the best models of human vision, and CNNs are currently the best performing models on this benchmark. More recently, \cite{Biscione2022} have demonstrated that CNNs can acquire a variety of visual invariance phenomena found in humans, namely, translation, scale, rotation, brightness, contrast, and to some extent, viewpoint invariance.

More relevant for the present context, \cite{hannagan-2021} demonstrated that training a biologically inspired, recurrent CNN (CORnet-Z; \citealt{kubilius-2018}) to recognise words lead the model to reproduce some key findings regarding visual-orthographic processing in the human visual word form area (VWFA) as observed with fMRI, such as case, font, and word length invariance. In addition, the model’s word recognition ability was mediated by a restricted set of reading-selective units. When these units were removed to simulate a lesion, it caused a reading-specific deficit, similar to the effects produced by lesions to the VWFA.

The current work explores this topic further by determining to what extent CNNs and ViTs account for human orthographic coding as measured through masked form priming effects. Given past reports of DNN-human similarity, it might be predicted that DNNs will account for some form priming effects. What is less clear is how well DNNs will capture orthographic similarity effects in comparison to various orthographic coding schemes and models of word identification specifically designed to explain form priming effects, among other findings. Given that DNNs do not include any hand-built orthographic knowledge and are trained to classify pixel images of words, it would be impressive if these models did as well. This would be particularly so given that we trained the models in a highly simplified manner that ignores many important features of how humans learn to identify words, as described below.

\section{Methods} \label{section methods}

\subsubsection{Human Priming Data}
Human priming data was sourced from the Form Priming Project (FPP; \citealt{adelman-2014}) and was used to assess how well various psychological DNN models account for orthographic similarity. FPP contains reaction times for 28 prime types across 420 six-letter word targets, gathered from over 924 participants. The prime types and priming effects are shown in Table \ref{tab:prime types}. To measure the priming effect size, the mean reaction time ($mRT$) of each prime condition is compared to the $mRT$ of unrelated arbitrary strings (e.g., ‘pljokv’ for the word ‘design’).


\begin{table}[!ht]
    \centering
    \resizebox{\textwidth}{!}{%
        \begin{tabular}{llll}
            \hline
            Prime   Type                         & Code Relative to 123456     & Prime Relative to `DESIGN' & Priming Score \\ \hline
            1: Identity                          & 123456                      & design               & 42.7          \\
            2: Final   deletion                  & 12345                       & desig                & 34.2          \\
            3: Suffix                            & 123456d                     & designl              & 33.7          \\
            4: Final   transposition             & 123465                      & desing               & 32.5          \\
            5: Medial   transposition            & 132456/124356/123546        & desgin               & 31.4          \\
            6: Medial   deletion                 & 13456/12456/12356/12346     & dsign                & 29.6          \\
            7: Final   substitution              & 12345d                      & desigj               & 29.5          \\
            8: Initial   substitution            & d23456                      & pesign               & 29.2          \\
            9: Initial   transposition           & 213456                      & edsign               & 29.0          \\
            10: Central   insertion              & 123d456                     & desrign              & 29.0          \\
            11: Prefix                           & d123456                     & mdesign              & 26.7          \\
            12: Half                             & 123/456                     & des                  & 25.8          \\
            13: Repeated letter                  & 123DD456                    & deshhign             & 25.5          \\
            14: Central-double-deletion          & 1256                        & degn                 & 24.9          \\
            15: Medial   substitution            & 1d3456/12d456/123d56/1234d6 & desihn               & 22.7          \\
            16: Neighbour   once removed         & 12d356/13d456/124d56/123d46 & dslign               & 21.8          \\
            17: 2 apart transposition            & 143256/125436               & degisn               & 20.2          \\
            18: Central   double insertion       & 123dd456                    & desaxign             & 19.4          \\
            19: All-transposed                   & 214365                      & edisng               & 16.8          \\
            20: Central   double substitution    & 12dd56                      & dewvgn               & 14.9          \\
            21: Reversed   halves                & 321654                      & sedngi               & 13.4          \\
            22: 3-apart-transposition            & 153426                      & dgsien               & 9.9           \\
            23: Interleaved   halves             & 415263                      & idgens               & 8.9           \\
            24: Transposed   halves              & 456123                      & igndes               & 8.8           \\
            25: Unrelated   pseudoword           & dddddd                      & voctal               & 4.8           \\
            26: Reversed   except initial        & 165432                      & dngise               & 2.9           \\
            27: Central   quadruple substitution & 1dddd6                      & dzbtkn               & 2.3           \\
            28: Unrelated   arbitrary            & dddddd                      & cbhaux               & 0             \\ \hline
        \end{tabular}%
    }
    \caption{The 28 prime types from the FPP. Each prime type denotes the transformation of a given target word (in this example the target word is DESIGN) into a string via transposing, removing, or adding letters. The numbers in the second column indicated the letters of the prime relative to the target. For example, the prime type `final-deletion' in the second row transforms the word `DESIGN' by deleting the last letter into the string `desig'. When multiple codes (e.g., 123/456) are specified, it indicates that each of these sub-conditions contains an equal number of targets. When `d' or `D' is specified, a random letter not found in the target is used. When `d' is specified multiple times, the same letter is not reused. When `D' is specified multiple times, the same letter is reused. The same transformations were applied to all targets. Adapted from \cite{adelman-2014}.}
    \label{tab:prime types}
\end{table}

\subsubsection{Orthographic coding schemes and word recognition models}
As discussed above, numerous orthographic coding schemes and word recognition models have been developed to account for orthographic priming effects in humans. Here we assess how well five coding schemes and three models of word identification account for the priming effects reported in the FPP dataset, namely:

\begin{enumerate}
    \item Orthographic coding schemes
          \begin{enumerate}
              \item Absolute position coding (used in \citealt{mcclelland-1981})
              \item Spatial Coding \citep{davis-2010A}
              \item Binary Open Bigram \citep{grainger-2004}
              \item Overlap Open Bigram \citep{gomez-2008}
              \item SERIOL Open Bigram \citep{whitney-2001}
          \end{enumerate}
    \item Full priming models
          \begin{enumerate}
              \item Interactive Activation Model \citep{mcclelland-1981}
              \item The Letters in Time  (LTRS) Model \citep{adelman-2011}
              \item The Spatial Coding Model \citep{davis-2001-solar}
          \end{enumerate}
\end{enumerate}

In order to determine the degree of priming we used the match value calculator created by \cite{davis-2010B} that implements the five orthographic coding schemes. For each coding scheme, the calculator takes two strings as input and returns a match value that indicates their predicted similarity. It is assumed that the greater the orthographic similarity, the greater the priming.  Note, for any given coding scheme and priming condition, the match value for any target word is the same across target words. For example, in the final deletion condition noted in Table 1, the example target given is DESIGN, but the same exact similarity score is computed for all the targets in this condition (because the prime and target all share the first 5 letters, with only the final letter mismatching).

For the models of visual word identification we assessed orthographic similarity on the basis of their predicted priming score. IA and SCM implementations were taken from \cite{davis-2010A} and LTRS model was taken from a simulator developed by \cite{adelman-2011} \footnote{http://www.adelmanlab.org/ltrs/}. For each model the $mRT$ was computed for each related prime condition and subtracted from the $mRT$ in the unrelated arbitrary condition to produce the priming score.

\subsubsection{DNN models}

We trained seven common convolutional networks (CNNs) and four Vision Transformer networks (ViTs). The convolutional models belong to the families of AlexNet \citep{krizhevsky-2012}, VGG \citep{simonyan-2014}, ResNet \citep{he-2016}, DenseNet \citep{huang-2016} and EfficientNet \citep{tan-2019}, and all Transformers were from the ViTs family \citep{dosovitskiy-2020}. All models were pretrained on ImageNet \citep{deng2009imagenet} to initialise the weights. The ViTs listed in Table \ref{tab:model performance} vary in their complexities and a number of properties including number of layers and the way that attention is implemented. 

After pretraining on ImageNet, the final classifier layer was removed and the models were trained to classify images of 1000 different words (the same number used by \citealt{hannagan-2021}), with each word represented locally.  All 420 six-letter words from From Priming Project were used and the remaining 580 were sourced from Google’s Trillion Word Corpus \citep{google-2011}. All words were presented in upper-case letters and the lengths of the 580 words were evenly distributed between three, four, five, seven, and eight letters, with 116 words chosen at each length. As with the Form Priming Project’s 420 words, the 580 words are chosen to not contain the same letter twice. All words were trained in parallel (there was no age-of-acquisition manipulation) and for the same number of trials (there was no frequency manipulation).  The complete list is available through the GitHub repository of the current study \footnote{https://github.com/Don-Yin/Orthographic-DNN}.


\begin{table}[!ht]
    \centering
    \resizebox{0.8\textwidth}{!}{
        \begin{tabular}{lllc}
            \hline
                                         & Family                  & Model           & \multicolumn{1}{l}{Accuracy(\%)} \\ \hline
            \multirow{7}{*}{Convolution} & AlexNet                 & AlexNet         & 99.3                             \\
                                         & DenseNet                & DenseNet169     & 99.8                             \\
                                         & EfficientNet            & EfficientNet-B1 & 99.9                             \\
                                         & \multirow{2}{*}{ResNet} & ResNet50        & 100                              \\
                                         &                         & ResNet101       & 99.9                             \\
                                         & \multirow{2}{*}{VGG}    & VGG16           & 99.7                             \\
                                         &                         & VGG19           & 99.8                             \\ \cline{1-2}
            \multirow{4}{*}{Transformer} & \multirow{4}{*}{ViT}    & ViT-B/16        & 98.2                             \\
                                         &                         & ViT-B/32        & 100                              \\
                                         &                         & ViT-L/16        & 99.9                             \\
                                         &                         & ViT-L/32        & 99.8                             \\ \hline
        \end{tabular}
    }
    \caption{DNN Models' performance on the word recognition task using the validation dataset. Accuracy denotes the probability that a model's prediction (the one with the highest probability) matches the correct response.}
    \label{tab:model performance}
\end{table}

\begin{figure}[!ht]
    \centering
    \begin{subfigure}[p]{\textwidth}
        \centering
        \adjustbox{scale=1,center}{
            \begin{tikzcd}[sep=tiny]
                \includegraphics[scale=0.30]{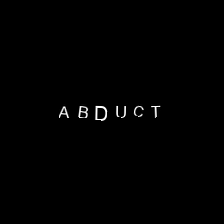} & \includegraphics[scale=0.30]{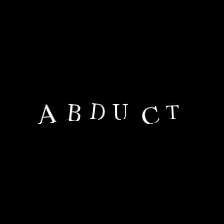} & \includegraphics[scale=0.30]{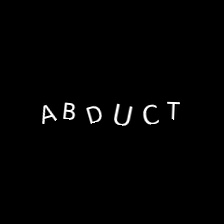} \\
                \includegraphics[scale=0.30]{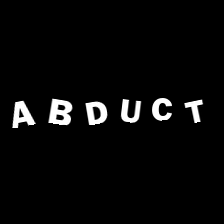} & \includegraphics[scale=0.30]{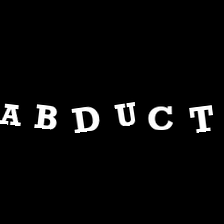} & \includegraphics[scale=0.30]{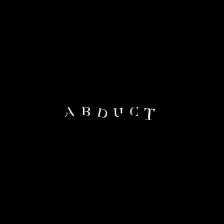} \\
                \includegraphics[scale=0.30]{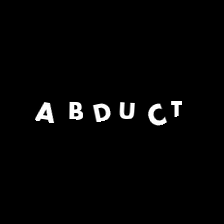} & \includegraphics[scale=0.30]{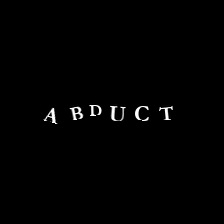} & \includegraphics[scale=0.30]{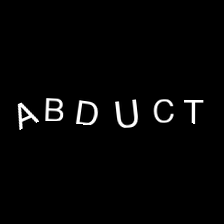}
            \end{tikzcd}
        }
        \caption{Training Data}

    \end{subfigure}\quad

    \begin{subfigure}[p]{\textwidth}
        \centering
        \adjustbox{scale=0.8,center}{
            \begin{tikzcd}[sep=0]
                \vcenter{\hbox{\includegraphics[scale=0.27]{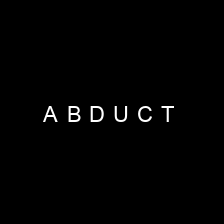}}} \arrow[r, bend left] \arrow[rr, bend left] \arrow[rrr, bend left] & \vcenter{\hbox{\includegraphics[scale=0.27]{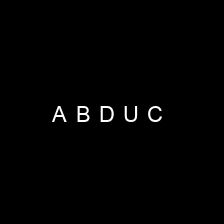}}} \arrow[l, bend right] & \vcenter{\hbox{\includegraphics[scale=0.27]{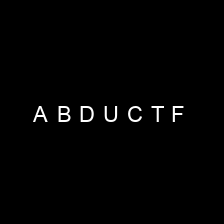}}} & \vcenter{\hbox{\includegraphics[scale=0.27]{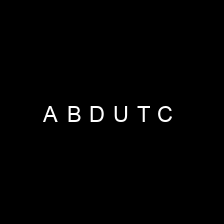}}} & ... \\
                \text{Target Stimulus}                                                                   & \text{Prime: Final Deletion}                       & \text{Prime: Suffix} & \text{Prime: Final Transposition} & ...
            \end{tikzcd}
        }
        \caption{Priming Data}
    \end{subfigure}\quad

    \caption{Image examples of \textbf{(a):} training data; \textbf{(b):} priming data using the word ABDUCT.}\label{fig: sample data}
\end{figure}

We employed data augmentation techniques to diversify the visual representation of each word by manipulating font, size, rotation of letters, and translation by changing the position of the image in space (See Figure \ref{fig: sample data} for some examples; for details of augmentation see Figure \ref{dig:process generation}). We generated 6,000 images for each word, resulting in a dataset containing 6,000,000 images. 5,000,000 images are used for training, while the remaining 1,000,000 are used for performance validation. The algorithm for generating datasets is described in detail in \ref{appendix: data generation}. For training the Adam optimizer and the cross-entropy loss function are used. A hyperparameter search was performed for the learning rate using a random grid-search, yielding a value of 1e-5. When the training average loss stops improving by a specified threshold of 0.0025, the training was terminated.  The accuracy of the models on the validation set of words is reported in Table \ref{tab:model performance}.

We then generated a dataset of prime words. To generate this dataset, each of the 420 Form Priming Project words was transformed into 28 prime types. For example, the target word ‘ABDUCT’ was transformed into ‘baduct’ for the ‘initial transposition’ condition, ‘abdutc’ for the ‘final transposition’ condition, and so forth. Each prime is used to generate an image using the Arial font, resulting in 11,760 images (420 target words $\times$ 28 prime types). No rotation or translation variations at the letter level are introduced, and all strings are positioned at the centre of the image using the same font size of 26, which is the average size used for training.

\subsection{Measuring orthographic similarity of the various DNNs}

To measure the the orthographic similarity between prime-target images we compared the unit activations at the penultimate layer using Cosine Similarity ($CS$) after each image was presented as an input to the network: 
\begin{equation}\label{eq:cosine similarity}
    CS(\textbf{A}, \textbf{B}):=\frac{\textbf{A} \cdot \textbf{B}}{\|\textbf{A}\|\|\textbf{B}\|}
\end{equation}

\noindent $CS$ ranges from -1 (opposite internal representation) to 1 (identical internal representation). We then computed the overall relation between human priming scores and model cosine similarity scores by calculating the correlation ($\tau$) between the mean human priming scores and the mean cosine similarities across conditions.  We also consider the relation between humans and models by exploring the (mis)match in priming and cosine similarity scores in the individual conditions, as discussed below.

We also included five baseline conditions to better understand the priming effects observed. First, the $CS$ between the pixels of the prime and target images was computed. This served as a baseline for determining the extent to which the models contribute to orthographic similarity scores beyond the pixel values between stimuli. In addition, we used ImageNet pretrained models (without any training on letter strings) and ImageNet pretrained models fine-tuned on 1000 classes of six-letter random strings for the CNNs and ViTs as four additional baselines. Rather than reporting tau for the individual models we report the average correlation tau across all the CNNs and ViTs, respectively. This was done to assess the role of training DNNs on English words on the pattern of form priming effects obtained. Using the aforementioned method, the mean cosine similarity score for each model condition was calculated, and the average correlation coefficient was used for each model class.

\section{Results}

\begin{figure}[!ht]
    \centering
    \includegraphics[width=1\linewidth]{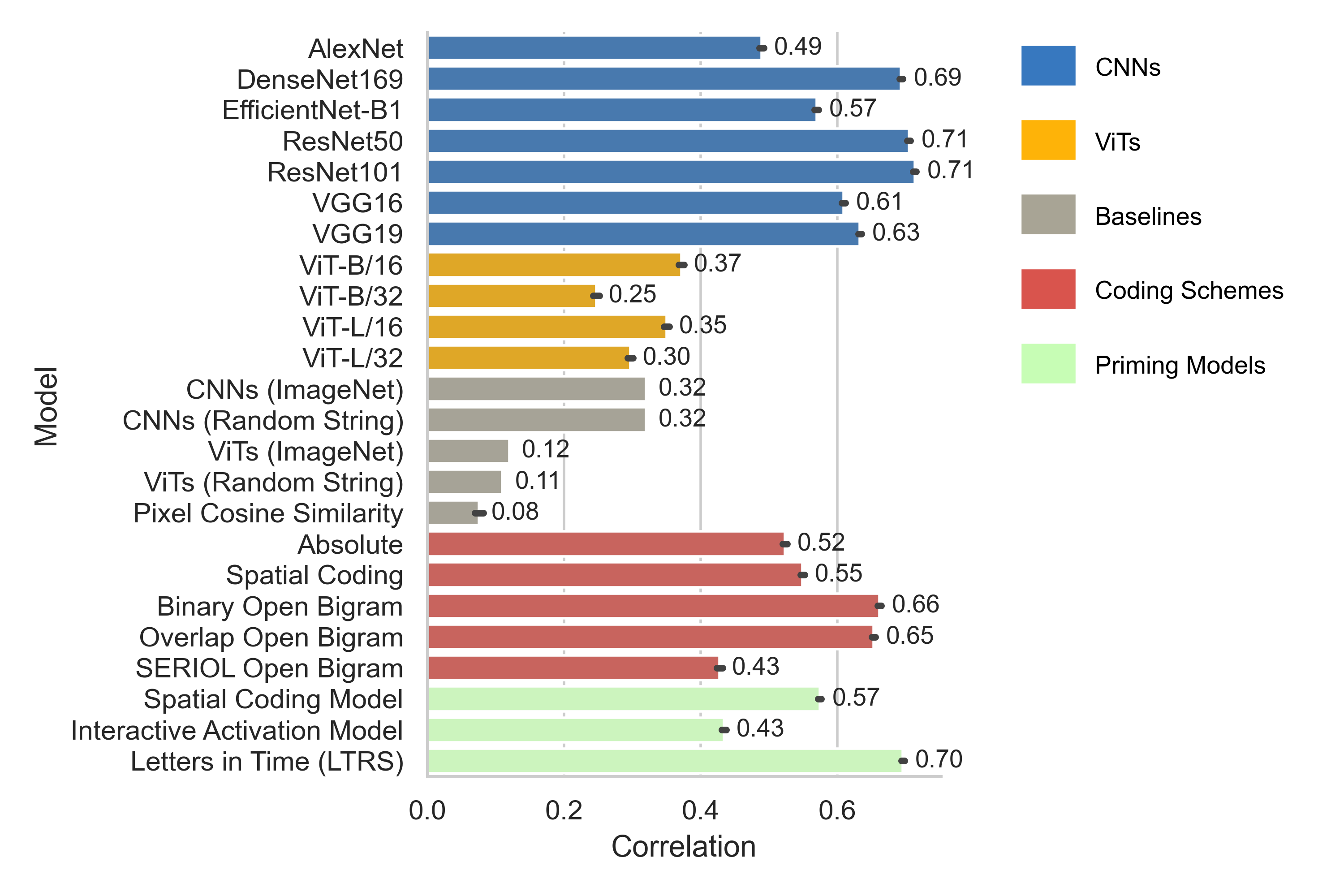}
    \caption{Priming (correlation) scores between model predictions and human data over the prime types for each DNN, coding scheme, priming model, and baseline. The term ``LTRS'' refers to the Letters in Time and Retinotopic Space. To obtain the standard error for each bar, we computed $tau$ 1,000 times by randomly sampling 28 mean cosine similarity scores with replacement across conditions. The error bar corresponds to the standard error of this vector.}
    \label{fig:priming scores}
\end{figure}

Figure \ref{fig:priming scores} plots Kendall’s correlation, over the 28 prime types, between the human priming data and the various orthographic coding schemes (as measured by match values), priming models (as measured by predicted priming scores), DNNs (as measured by cosine similarity scores), as well as various baseline measures of similarity.

The most striking finding is that the CNNs did a good job in predicting the pattern of human priming scores across conditions, with correlations ranging from $\tau$ = .49 (AlexNet) to $\tau$ = .71. (ResNet101) with all p-values $<$ .01. Indeed, the CNNs performed similarly to the various orthographic coding schemes and word recognition models, and often better.  This contrasts with the relatively poor performance of the Transformer networks, with $\tau$ ranging from .25 to .38.

Importantly, the good performance of the CNNs was not due to the pixel value similarity between the prime-target images, as the pixel control condition (pixCS) has no significant correlation with the human priming data. It is also not simply the product of the architectures of the CNNs, as the predictions were much poorer for the CNNs that were pretrained on ImageNet but not trained on English words. Rather, it is the combination of the CNN architectures with training on English words that led to good performance. For a complete set of correlations between human priming data, DNNs' cosine similarity scores, orthographic coding similarity scores, and priming scores in psychological models, see \ref{appendix: data generation}, Figure \ref{fig:correlation main}.

A more detailed assessment of the overlap between DNNs, orthographic coding schemes, and word identification models is provided in Figures \ref{fig:ridge NNs} and \ref{fig:ridge psychological models} that depict the distribution of responses in each condition for all models as well as summarise the priming results per condition. From this, it is clear that all the CNNs had particular difficulty in predicting the priming in `half' condition (as indicated by the red arrow) in which either the first three letters or the final three letters served as primes (the CNNs substantially underestimated the priming in this condition).  A similar difficulty was found in many of the psychological models as well, but the effect was not quite so striking. There were no other prime conditions that led to such a large and consistent error in any model or coding scheme.

One surprising result from the form priming project was that there was little evidence that external letters were more important than internal letters. For instance, final and initial substitutions produced more priming than medial substitutions, and similar priming effects were obtained for final, medial, and initial transpositions, with slightly less priming for initial transpositions. This contrasts with the common claim that external letters are more important than medial letters for visual word identification \citep{Estes1976}, although the past evidence for this in masked priming is somewhat mixed (e.g., \citealt{Perea2003}). Interestingly, most of the CNNs and ViTs showed similar effects across the three substitution conditions, and slightly less priming in the initial transposition condition, and thus also predicted little extra importance attributed to external letters.

\begin{landscape}
    \def\names{{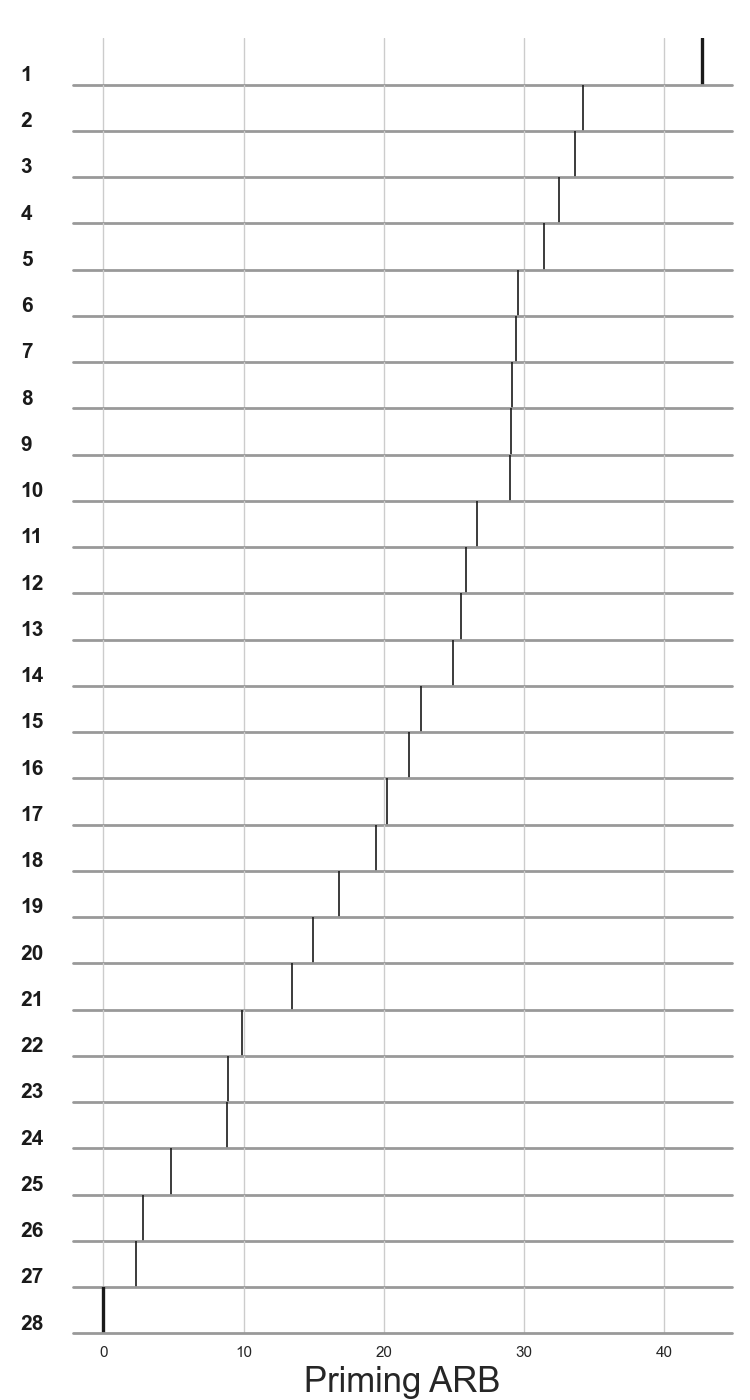/Priming},{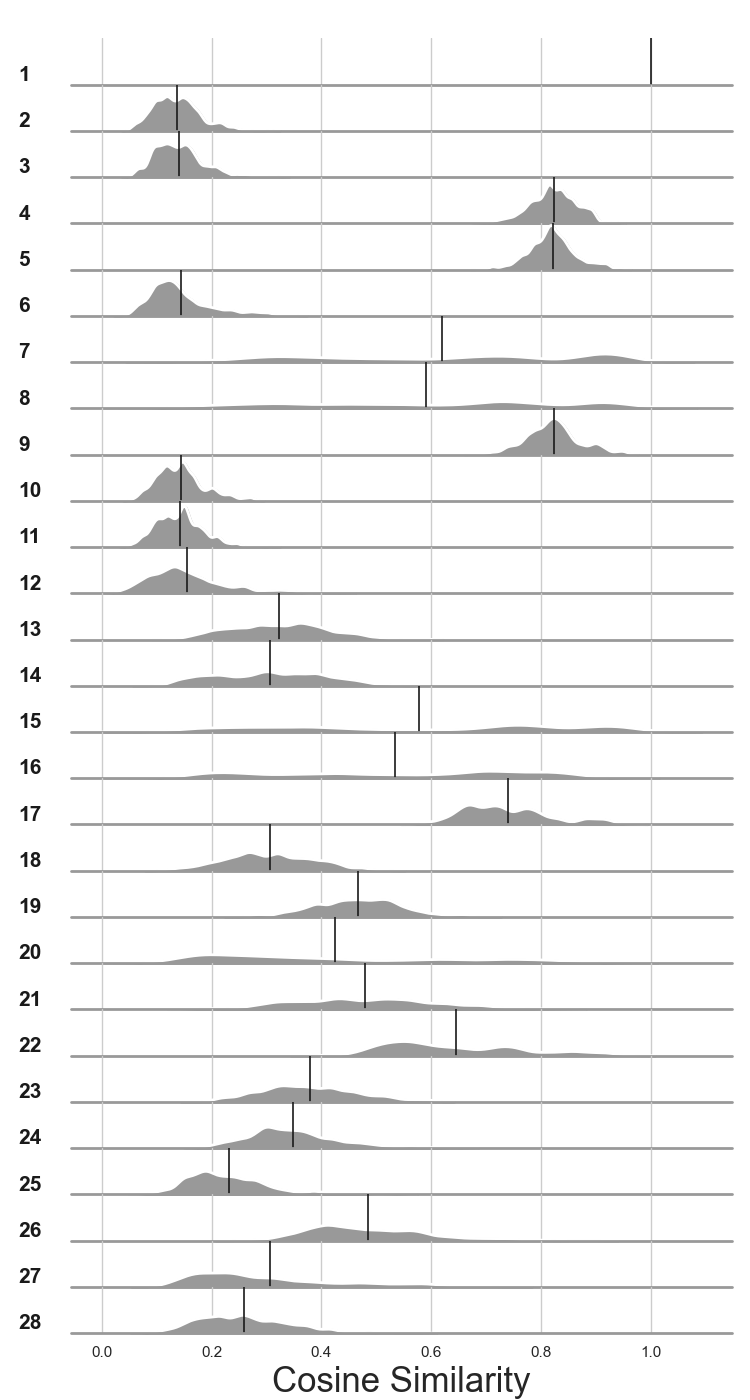/pixCS}, {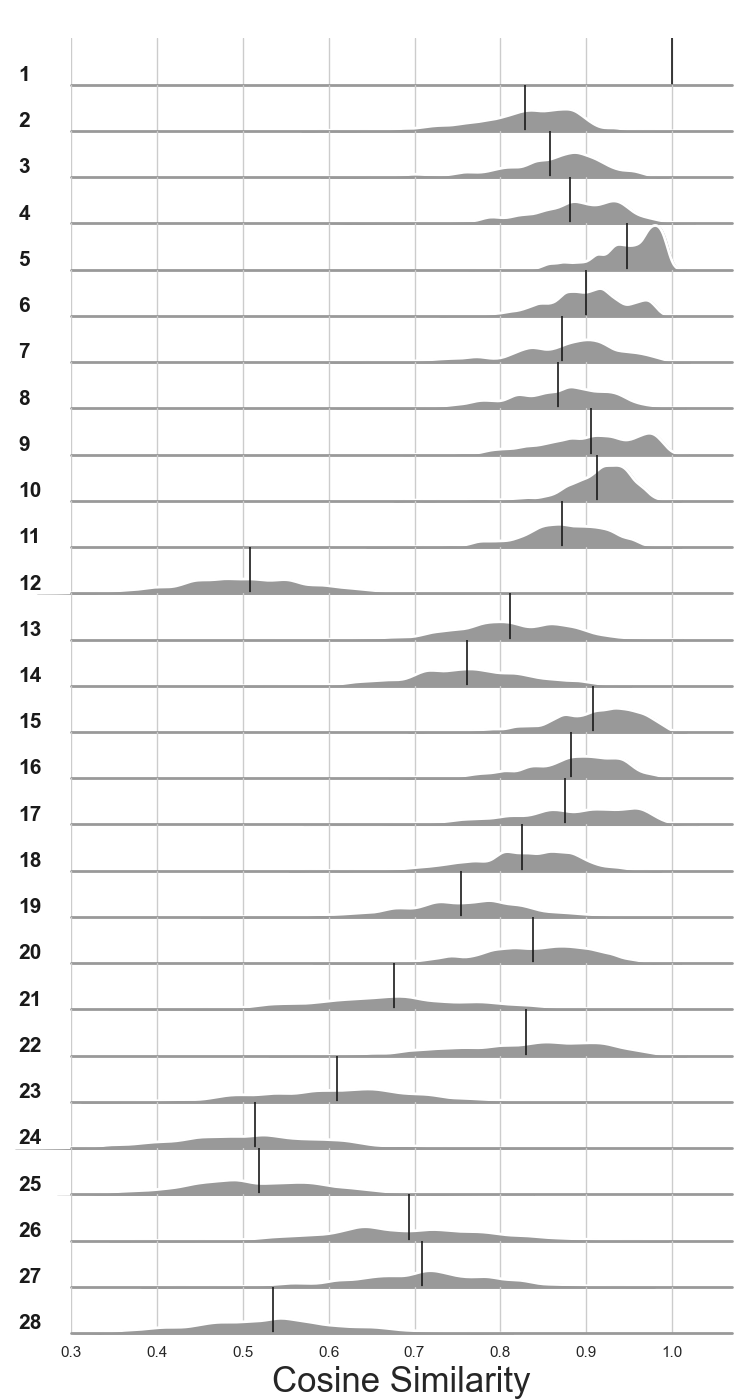/AlexNet}, {densenet169/DenseNet169},{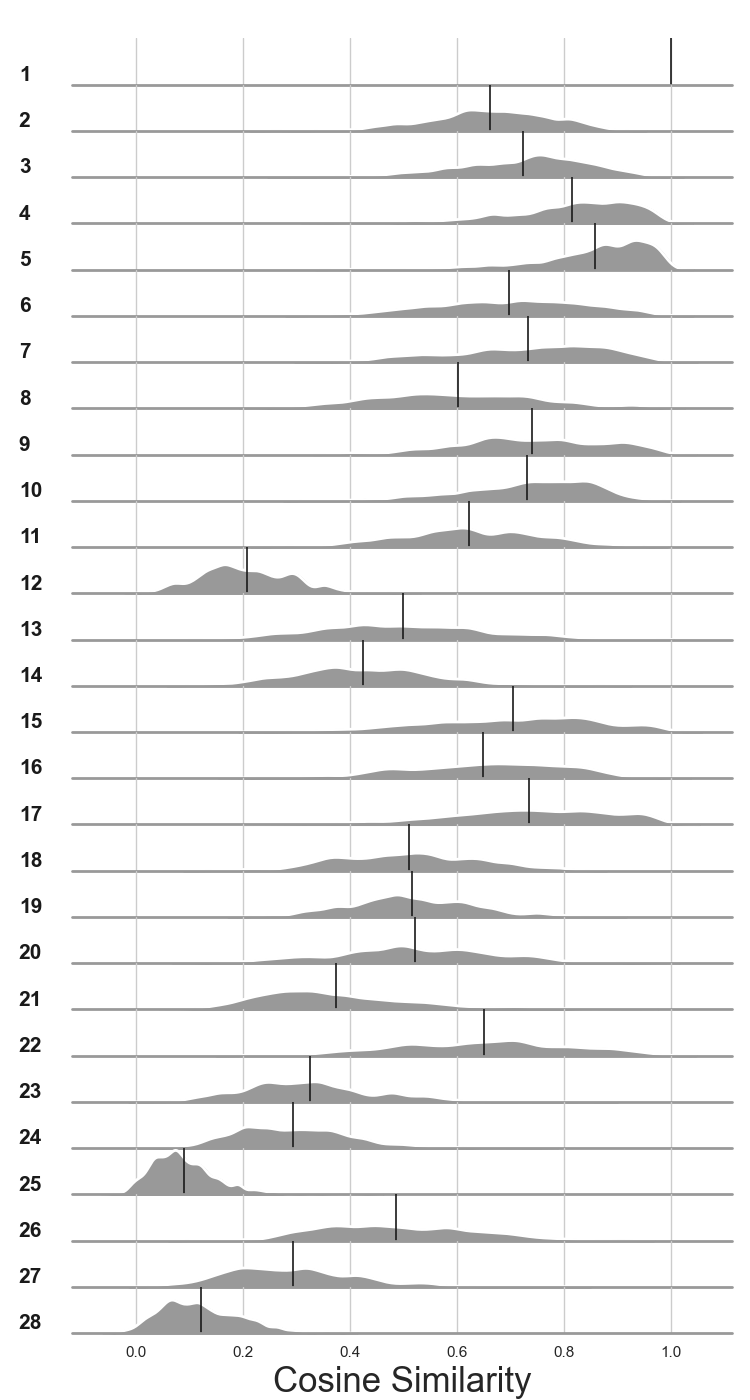/EfficientNet}, {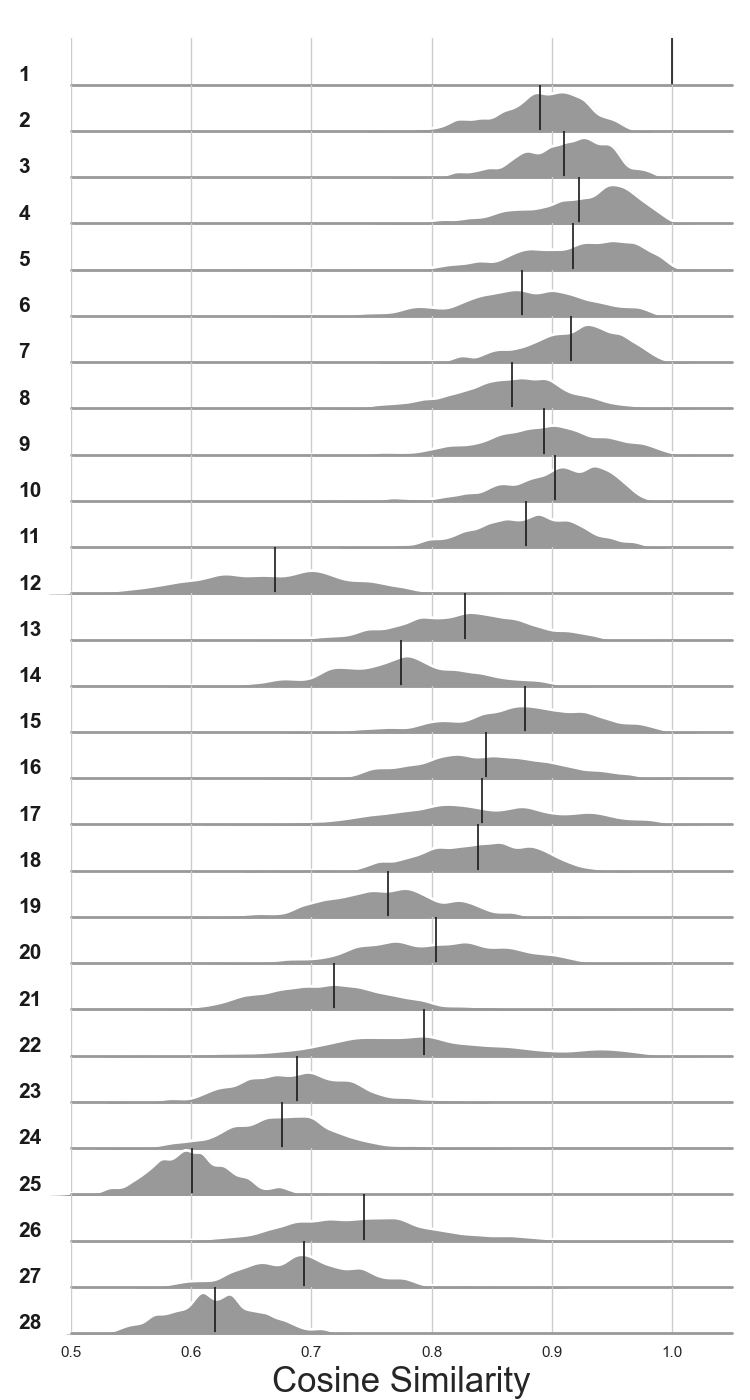/ResNet50}, {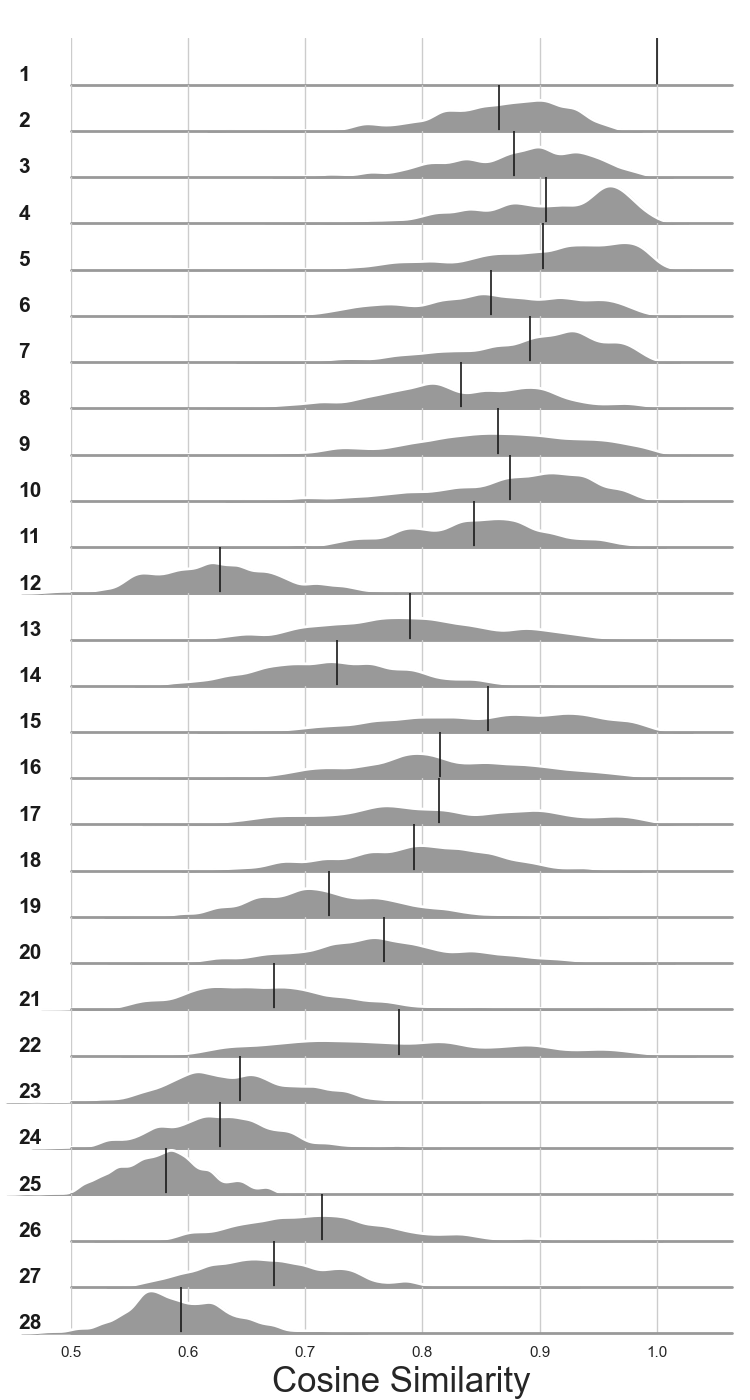/ResNet101}, {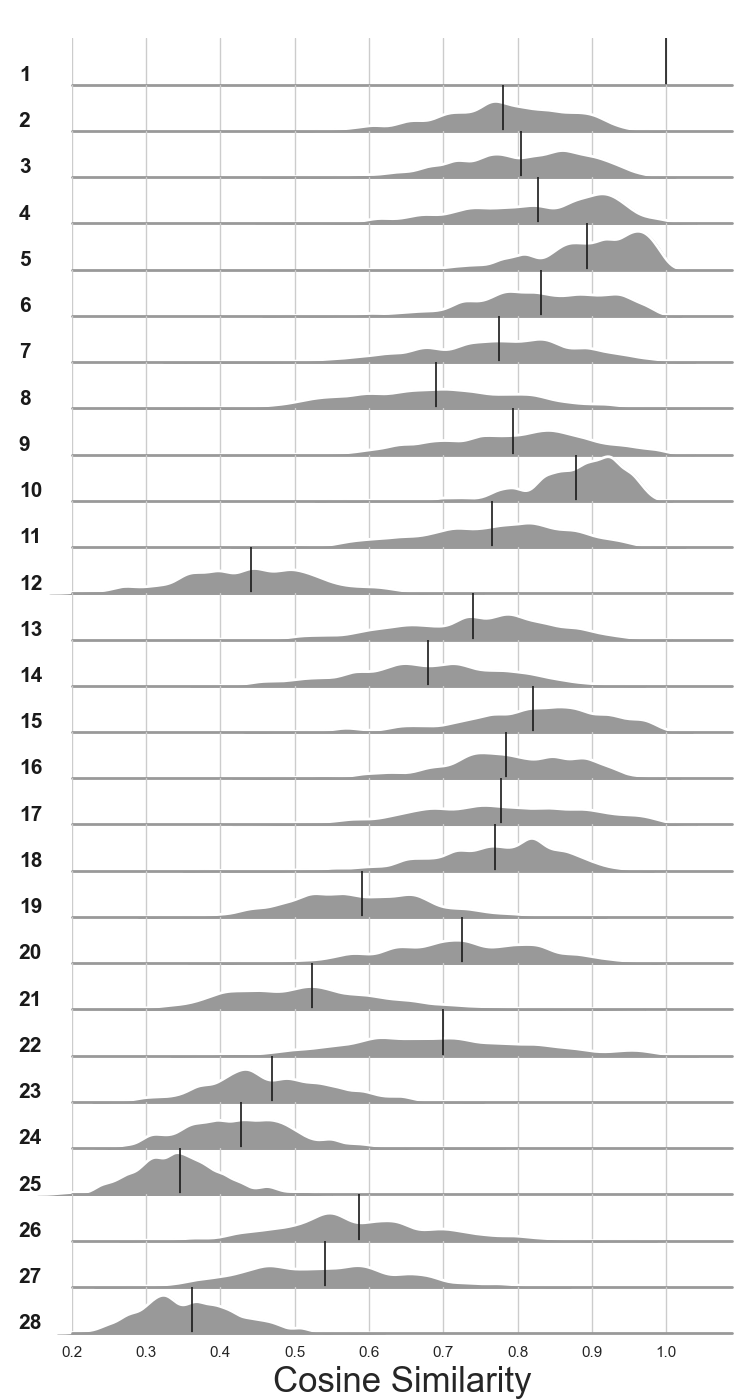/VGG16}, {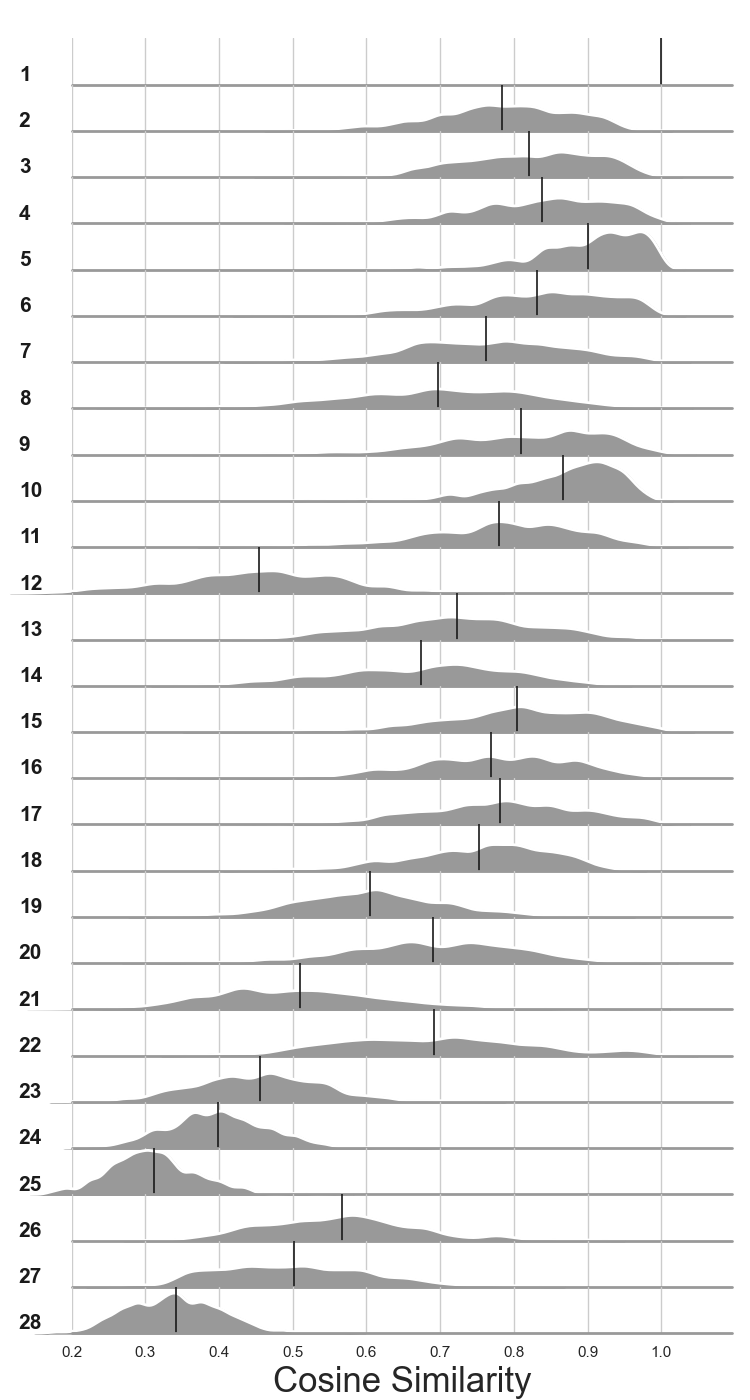/VGG19}, {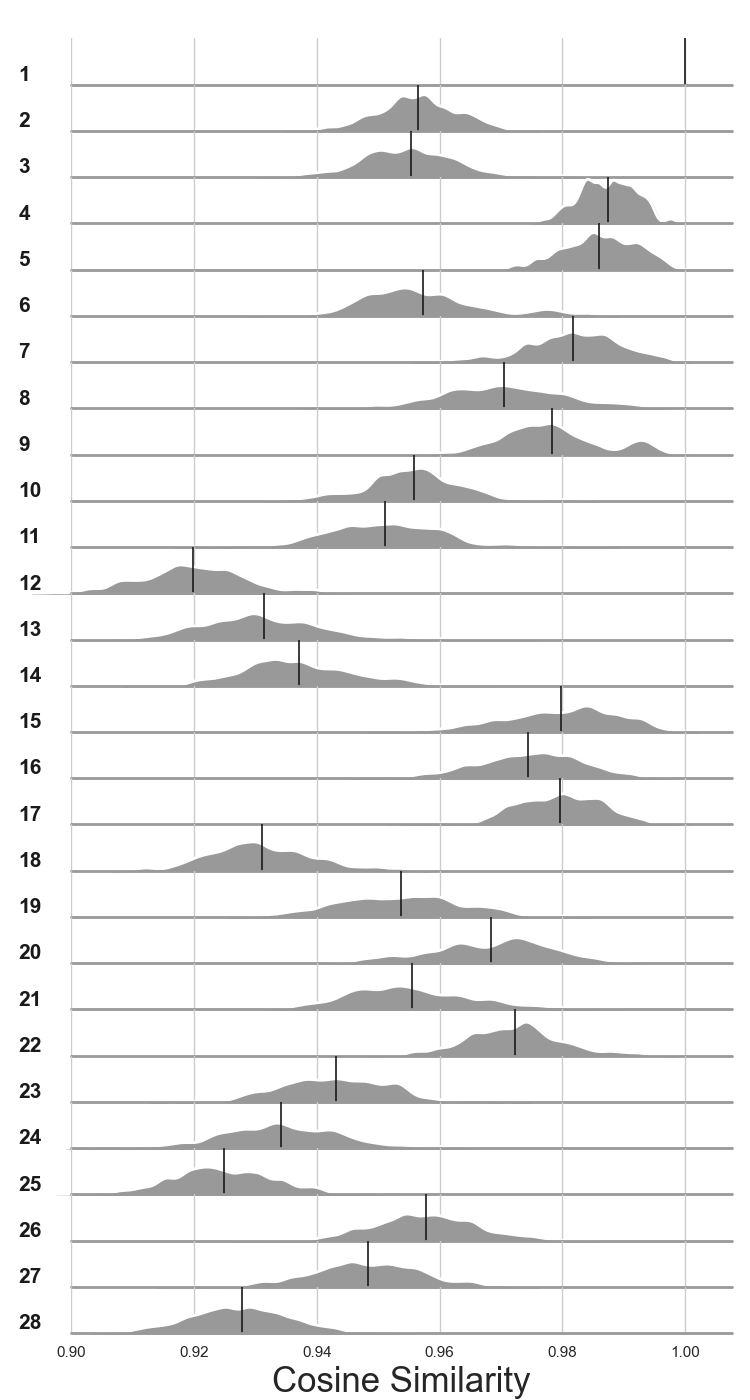/ViT-B16}, {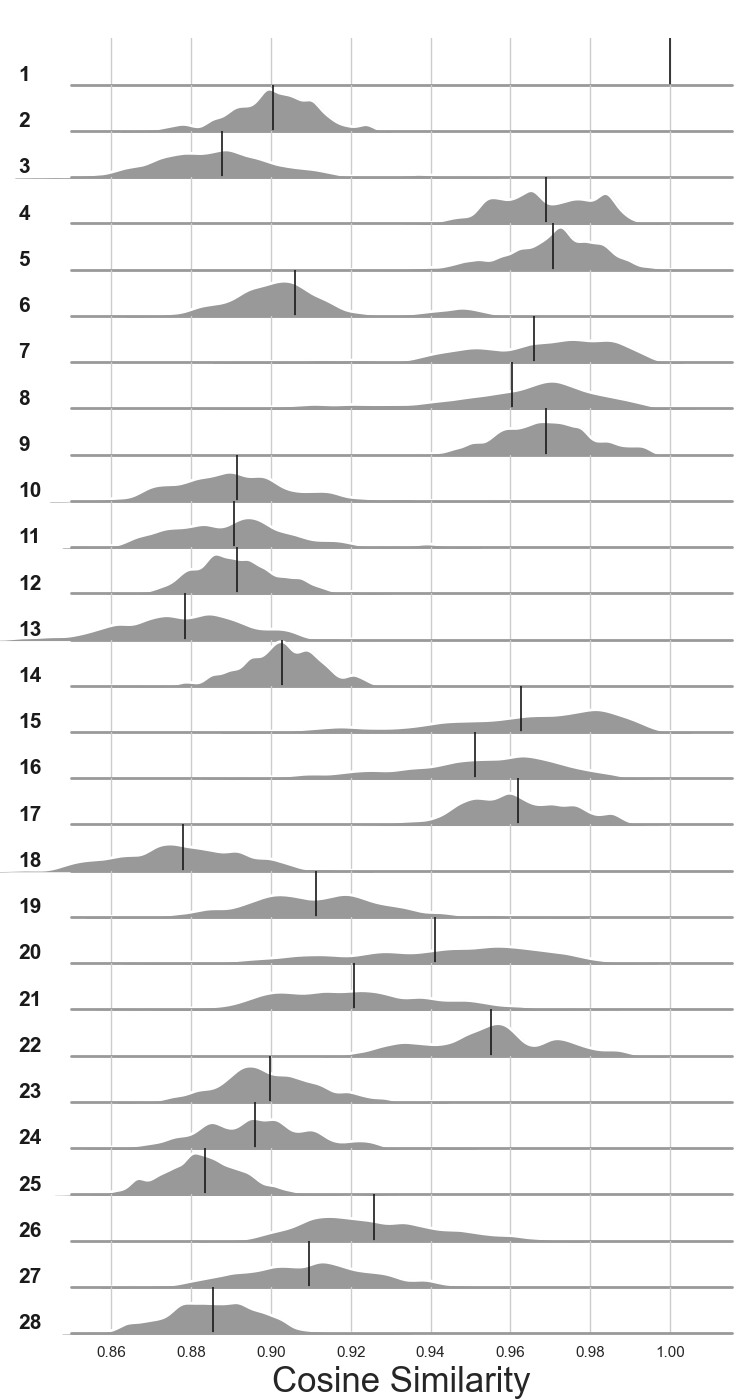/ViT-B32}, {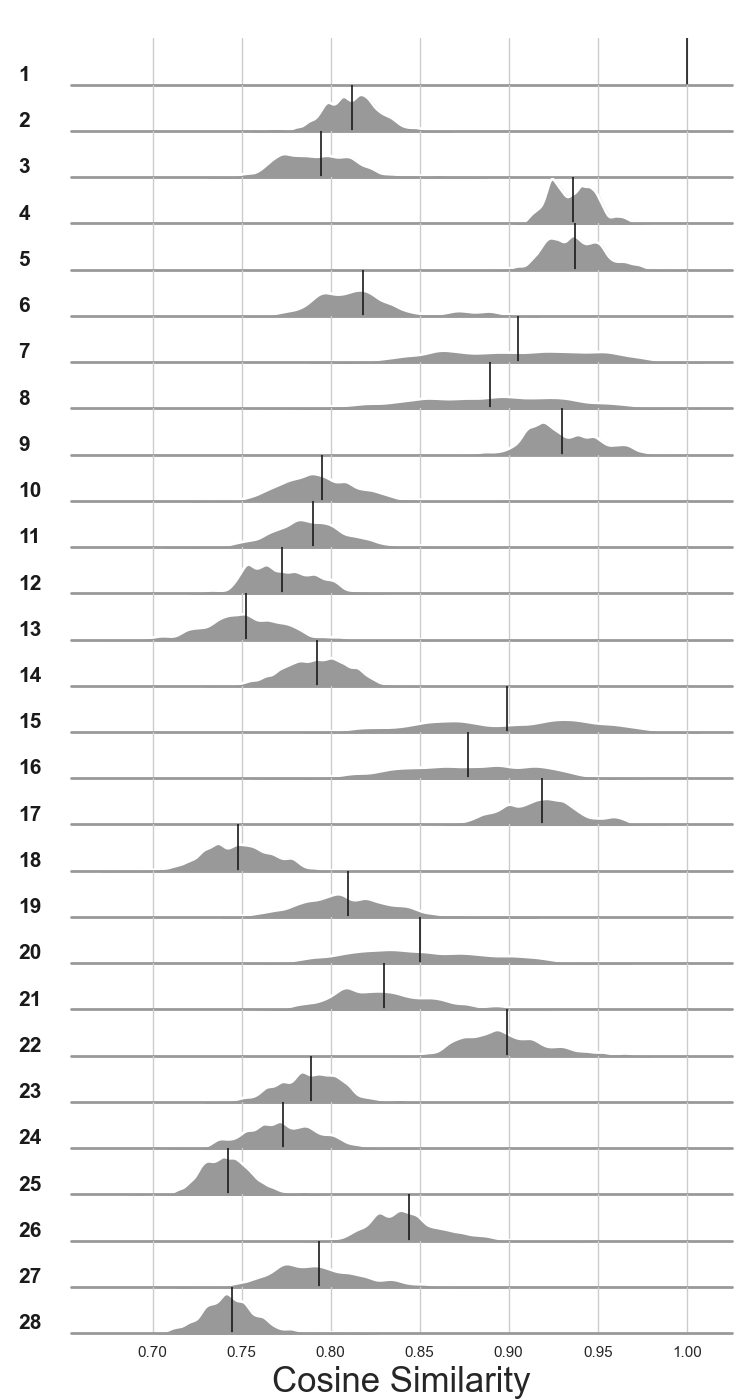/ViT-L16}, {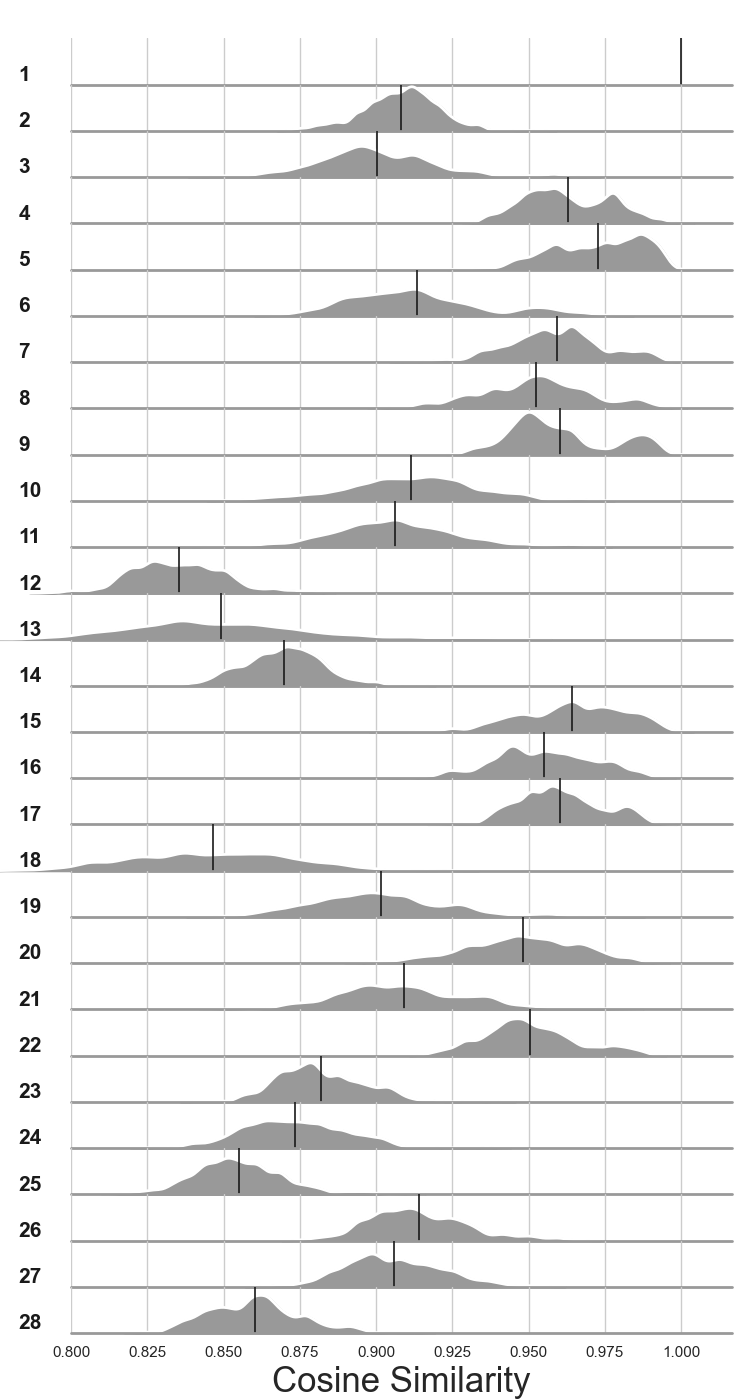/ViT-L32}}

    \begin{figure}[!ht]
        \centering
        \foreach \name/\subcap in \names {%
            \begin{subfigure}[p]{0.143\textwidth}
                \includegraphics[width=1.23\linewidth]{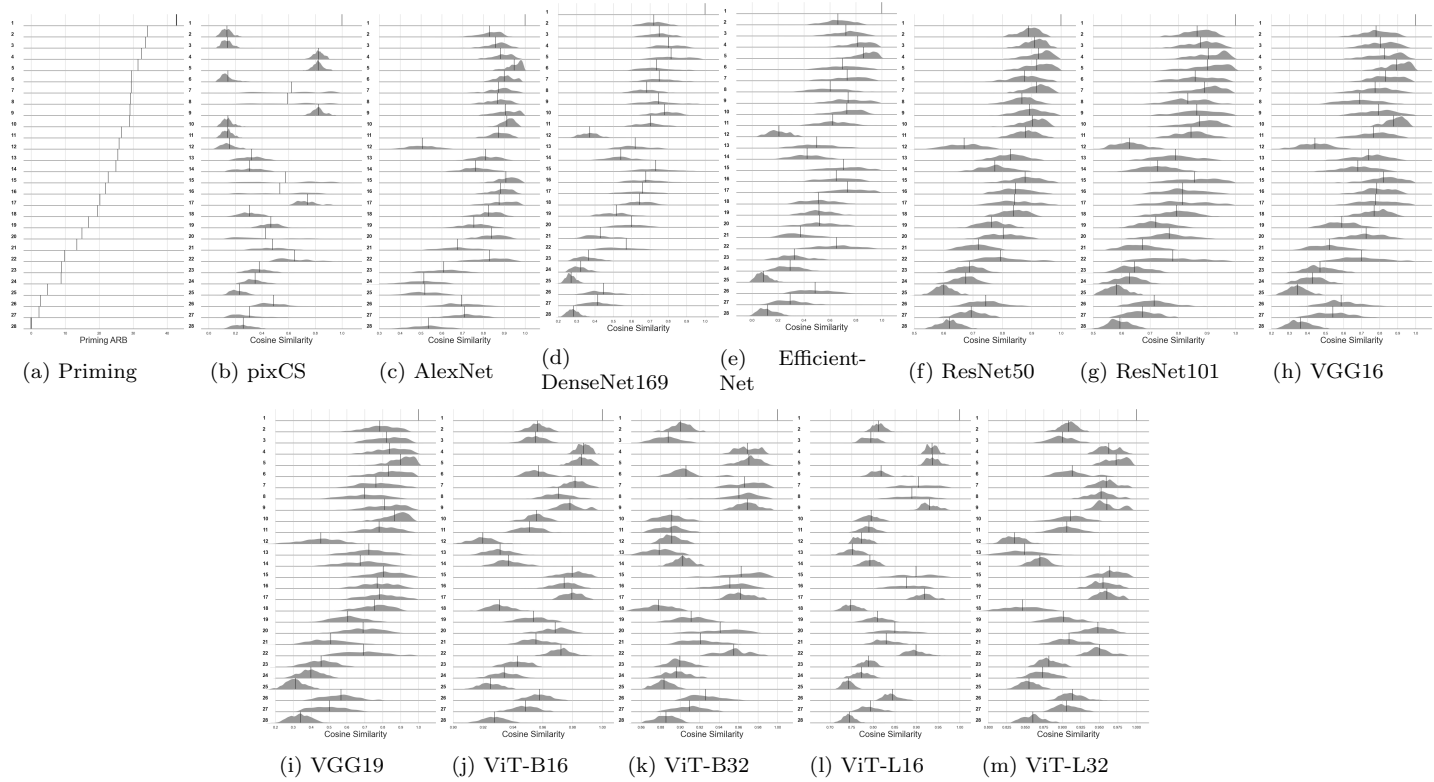}
                \caption{\subcap}
            \end{subfigure}\quad
        }
        \caption{Distributions of Human Priming and the DNN Models Perception. For each subplot, the x-axis is the metric specific similarity measure and the y-axis is the prime condition. Prime types are ordered according to the size of priming effect in the human data (largest to smallest), as in Table \ref{tab:prime types}. As illustrated in the first row, the `identity' condition has the strongest priming effect as it has the highest priming score of 42.69ms. Priming score for a condition is the difference of its mRT and the `unrelated arbitrary' condition. See table \ref{tab:prime types} for the index of the 28 conditions.}\label{fig:ridge NNs}
    \end{figure}
\end{landscape}

\begin{landscape}
    \def\names{{human_data_ridge_plot_priming_arb.png/Priming},{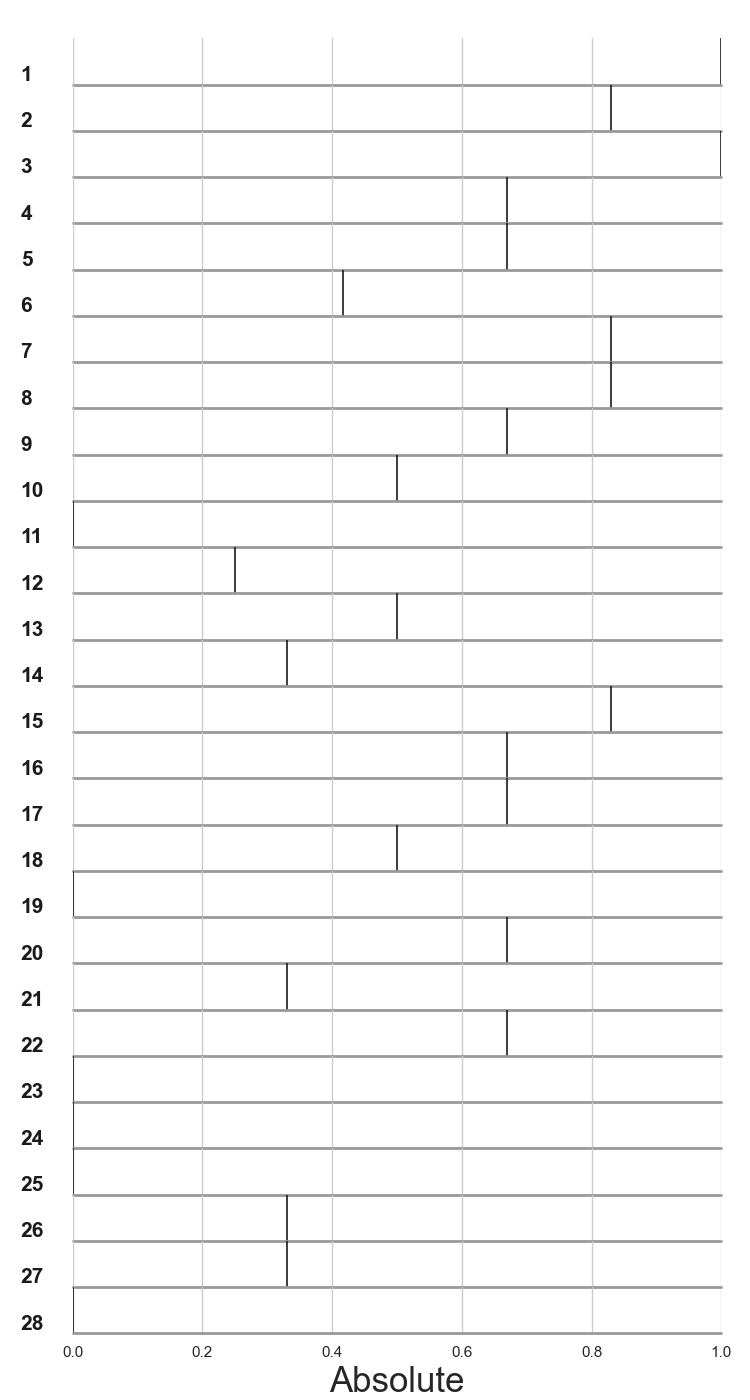/Absolute},{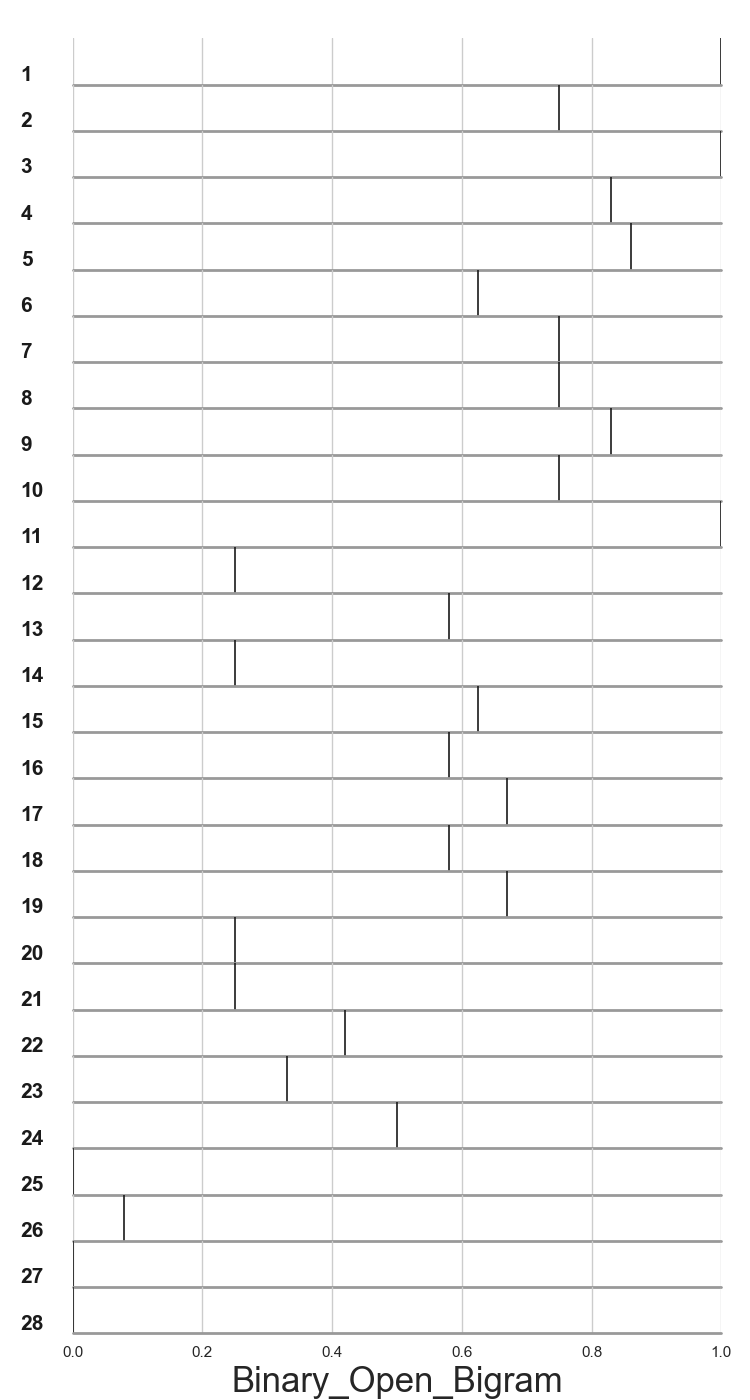/Binary Open Bigram}, {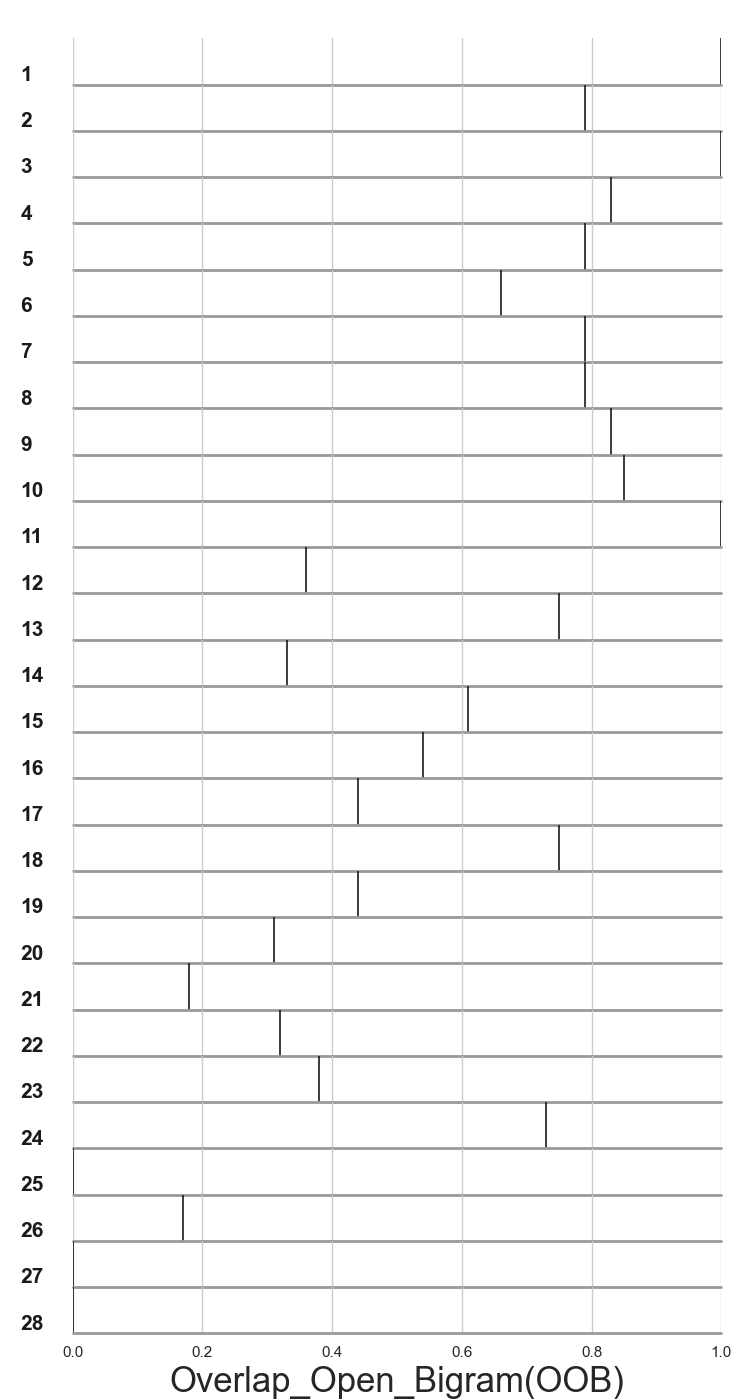/Overlap Open Bigram}, {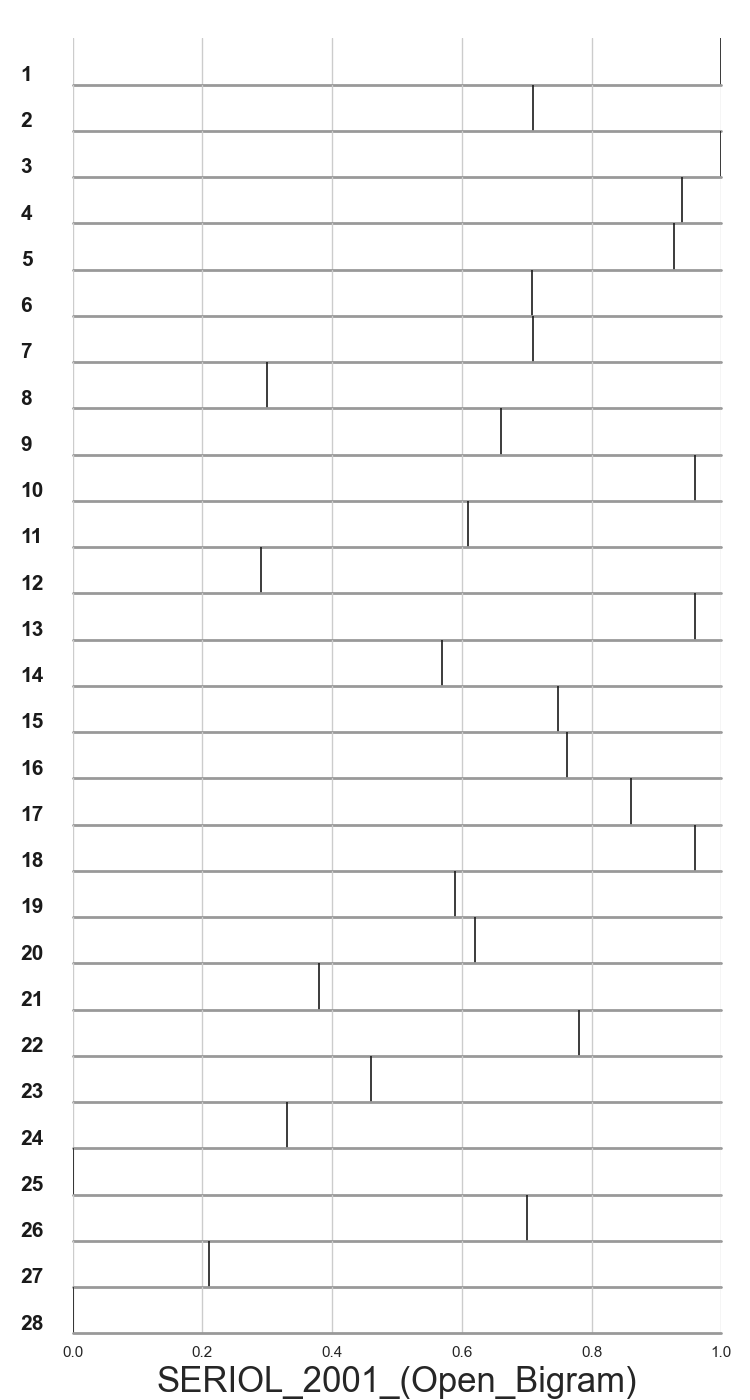/SERIOL Open Bigram}, {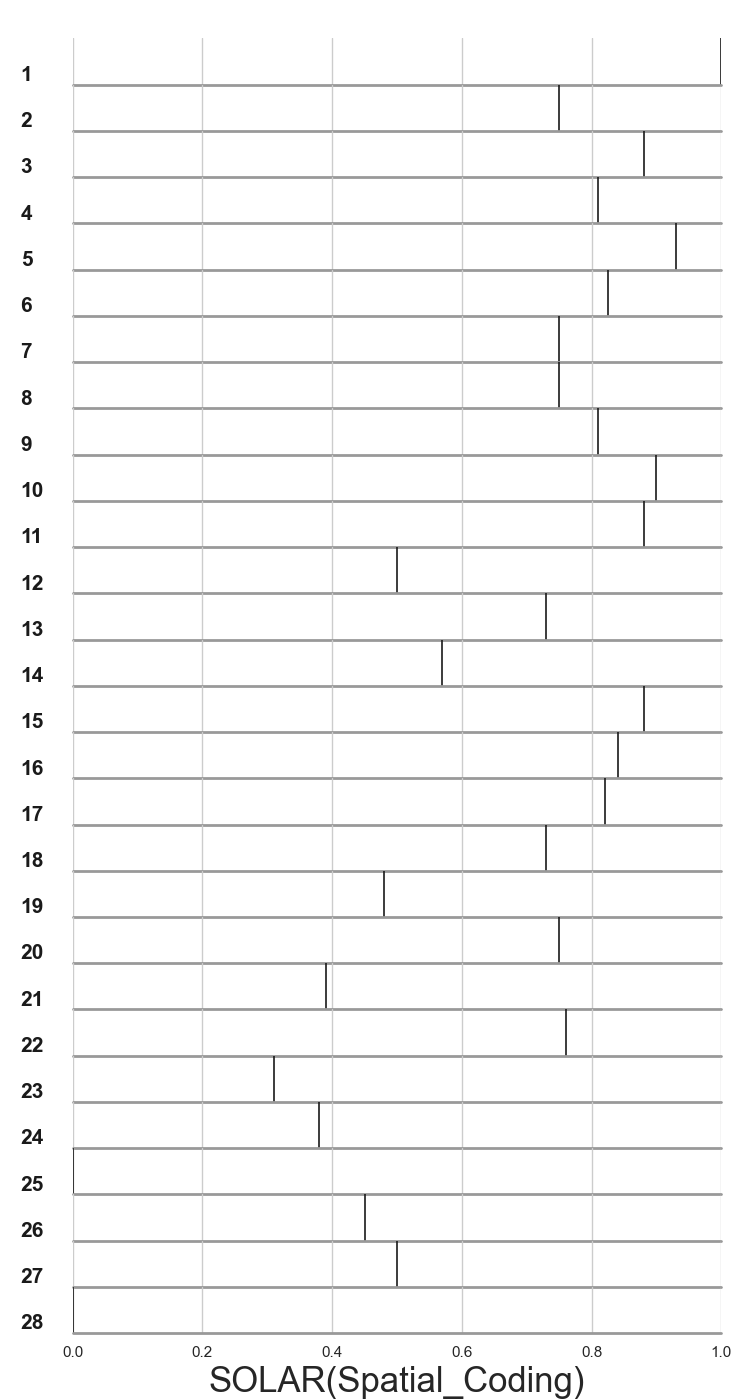/Spatial Coding Model}, {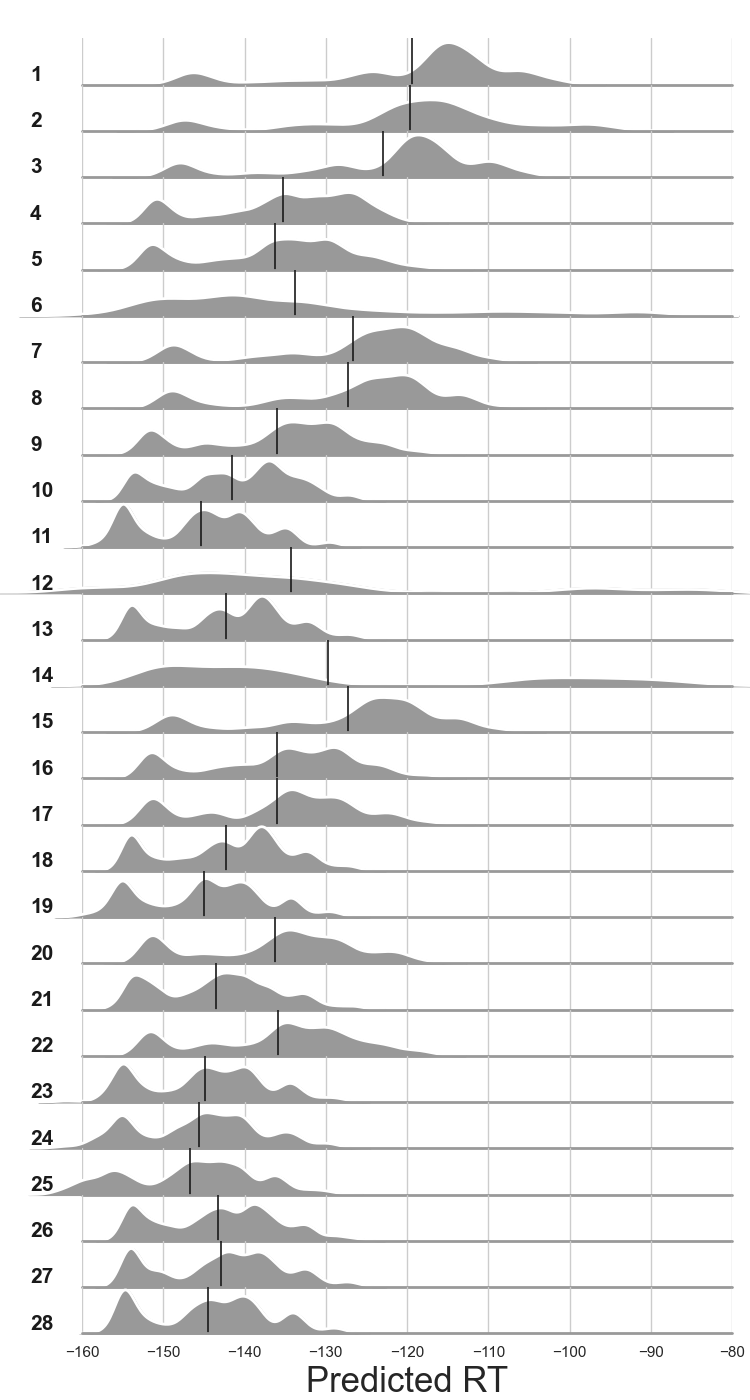/IA},{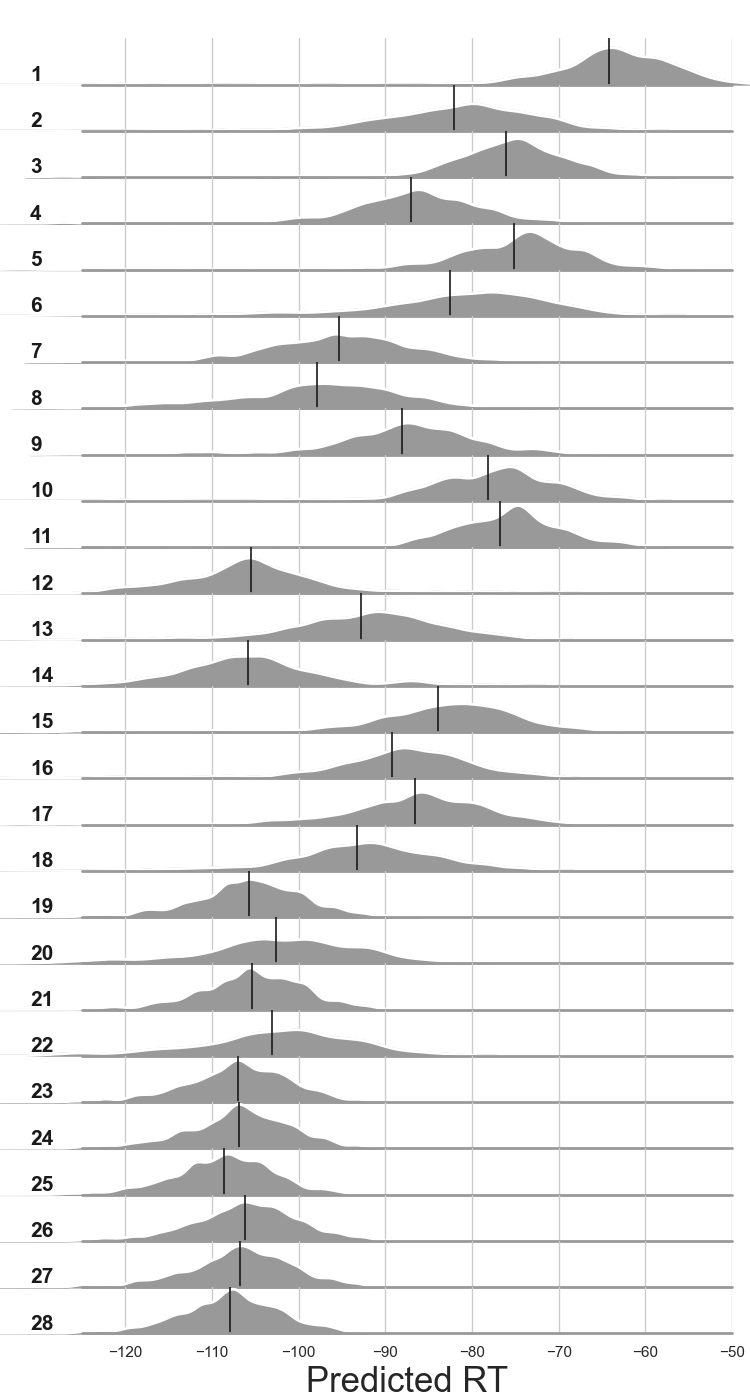/Spatial Coding}, {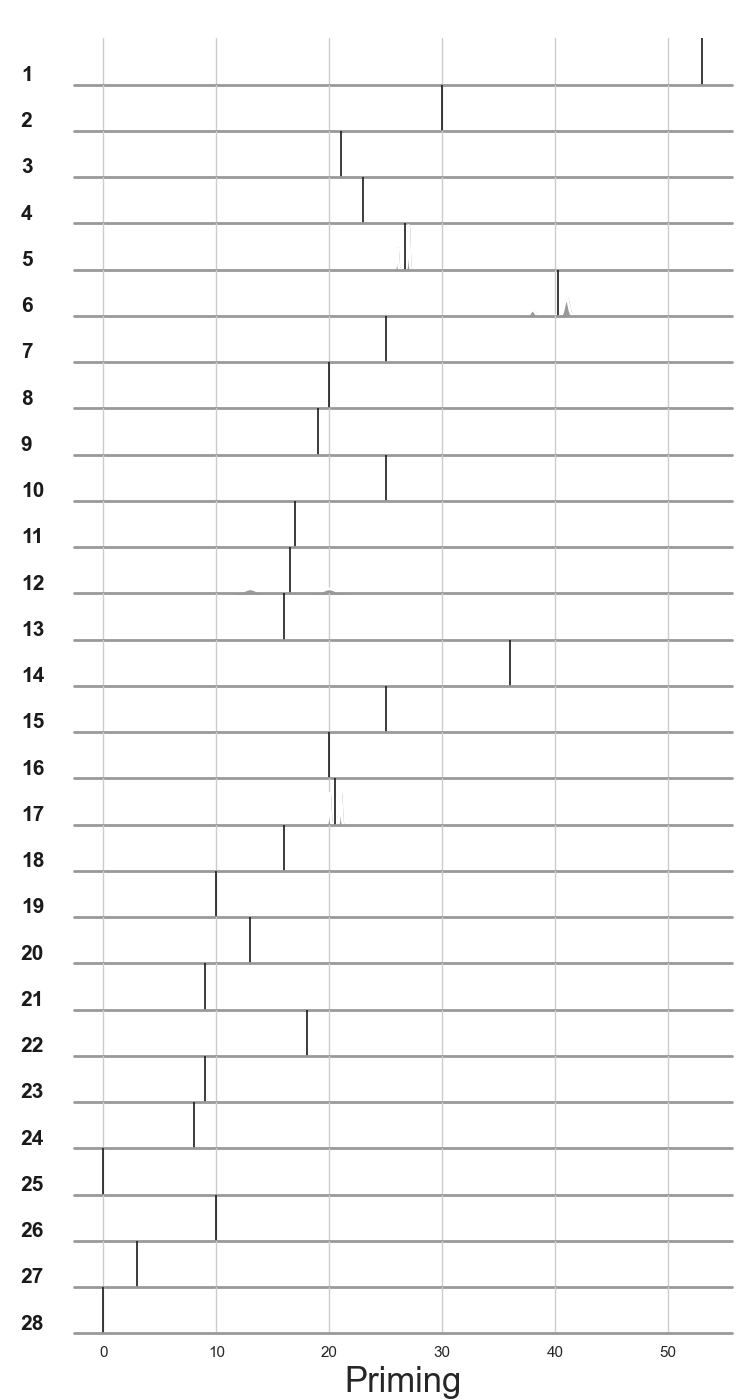/LTRS}}

    \begin{figure}[!ht]
        \centering
        \foreach \name/\subcap in \names {%
            \begin{subfigure}[p]{0.19\textwidth}
                \includegraphics[width=\linewidth]{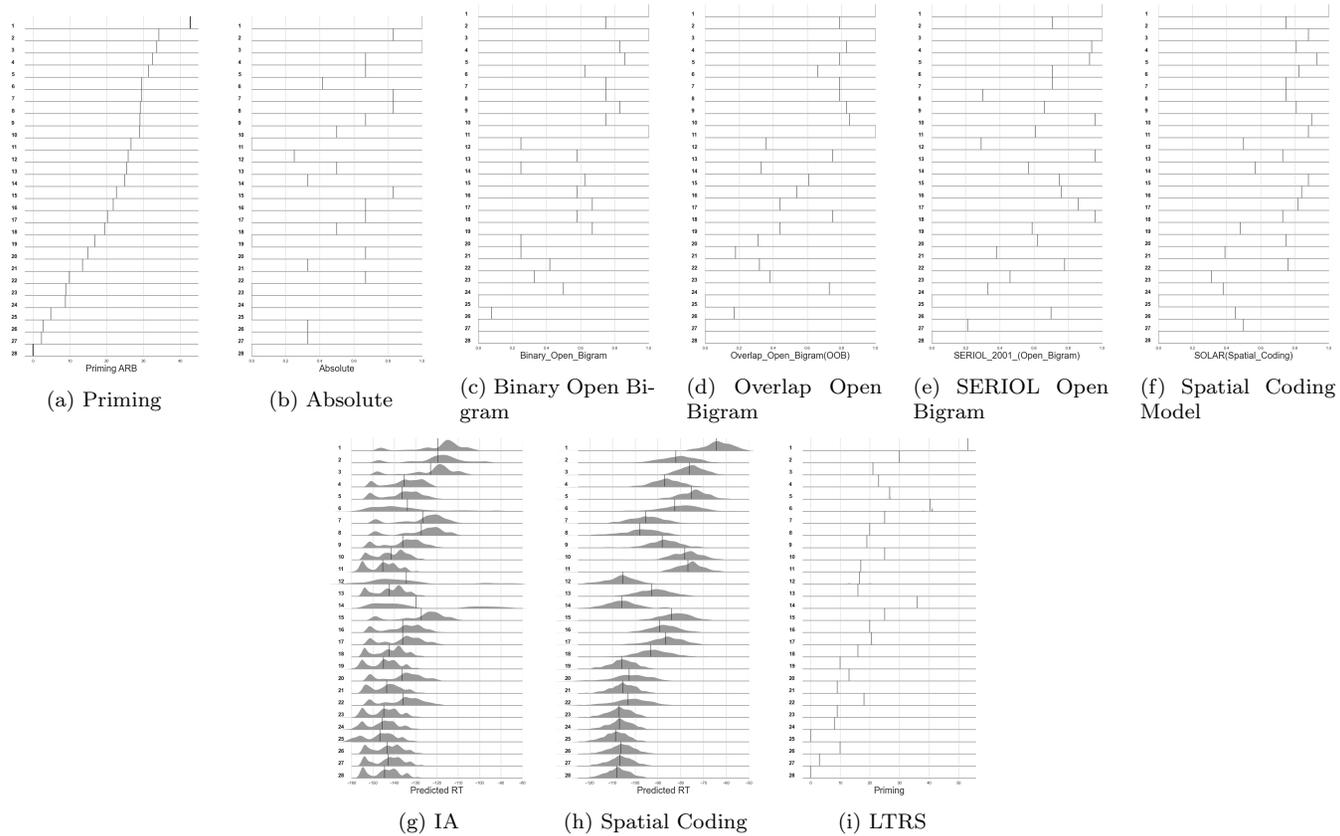}
                \caption{\subcap}
            \end{subfigure}\quad
        }
        \caption{Distributions of Human Priming and the coding schemes. For each subplot, the x-axis is the metric specific similarity measure and the y-axis is the prime conditions. See table \ref{tab:prime types} for the index of the 28 conditions. The Interactive Activation and Spatial Coding Model values are made negative to obtain a positive correlation coefficient with human priming data, as they represent estimated $RT$s that is negatively correlated with priming effect size.}\label{fig:ridge psychological models}
    \end{figure}
\end{landscape}

\begin{table}[h!]
\centering
\scalebox{1}{
\begin{tabular}{lll|lll}
\hline
Model           & condition SN & $r$  & Model    & condition SN & $r$  \\ \hline
AlexNet         & F            & .53 & VGG-19   & F            & .47 \\
                & I            & .53 &          & I            & .31 \\
                & M            & .52 &          & M            & .49 \\ \hline
DenseNet-169    & F            & .56 & ViT-B/16 & F            & .41 \\
                & I            & .39 &          & I            & .6  \\
                & M            & .61 &          & M            & .29 \\ \hline
EfficientNet-B1 & F            & .52 & ViT-B/32 & F            & .32 \\
                & I            & .55 &          & I            & .36 \\
                & M            & .58 &          & M            & .2  \\ \hline
ResNet-101      & F            & .51 & ViT-L/16 & F            & .32 \\
                & I            & .38 &          & I            & .38 \\
                & M            & .56 &          & M            & .21 \\ \hline
ResNet-50       & F            & .55 & ViT-L/32 & F            & .54 \\
                & I            & .37 &          & I            & .57 \\
                & M            & .56 &          & M            & .31 \\ \hline
VGG-16          & F            & .55 &          &              &      \\
                & I            & .45 &          &              &      \\
                & M            & .54 &          &              &      \\ \hline
\end{tabular}}
\caption{Pearson correlation coefficients between visual similarity ratings of upper case letters and cosine similarity scores observed in the initial, middle, and final substitution conditions for various DNN models. The SN (substitution) conditions I, M, and F denote the initial, middle, and final substitution, respectively. All p values $<$ .001.}\label{tab:visual sim}
\end{table}

A key feature of all the psychological models is that they code letters in an abstract format such that there is no variation in visual similarity between letters (different letters are simply unrelated in their visual form). This manifests itself in the fact that there is no distribution in priming scores in each of the 28 prime conditions for the orthographic priming schemes and the LTRS model. For the IA and Spatial Coding model there is variation in priming score in each condition, but this reflects the impact of lexical access in the models (e.g., the impact of word frequency or lexical competition) rather than any influence of visual similarity.

By contrast, in the case of the CNNs and ViTs, the input is an image in pixel space, and accordingly, it is possible that the visual similarity is contributing to the distribution of priming scores observed in each of the priming conditions. To test for this, we obtained human visual similarity ratings between upper case letters \citep{Simpson2012} and assessed whether these scores correlate with the cosine similarity scores observed in the initial, middle, and final substitution conditions. In these conditions, the visual similarity of all the letters is the same other than the substituted letter, and the question is whether the similarity score computed with the model correlates with the similarity scores produced by humans. As can be seen Table \ref{tab:visual sim}, there was a strong correlation, for all models in all the letter substitution conditions. This is an advantage of DNNs over current psychological models given that masked priming in humans is also sensitive to visual similarity of letter transpositions \citep{kinoshita-2013B, Forster1987-visual-sim, Perea2008}. 






\section{Discussion}

A wide number of orthographic coding schemes and models of visual word identification have been designed to account for masked form priming effects that provide a measure of the orthographic similarities between letter strings. Here, we assessed how well these standard approaches account for the form priming effects reported in the Form Priming Project \citep{adelman-2014} and compared the results to two different classes of DNNs (CNNs and Transformers). Strikingly, the CNNs we tested did similarly, and in some cases better, than the psychological models specifically designed to explain form priming effects.  This is despite the fact that the CNN architectures were designed to perform well in object identification rather than word identification, despite the fact that the models were trained to classify pixel images of words rather than hand-built and artificially encoded letter strings, and despite the fact that the models were trained to classify 1000 words in a highly simplified manner.

By contrast, we found that visual transformers are less successful in accounting for form priming effects in humans, suggesting that these models are identifying images of words in a non-human like way. Still, both CNNs and Transformers did better than psychological models in one important respect, namely, they can account for the impact of the visual similarity of primes and targets on masked priming. By contrast, all current psychological models cannot given that their input coding schemes treat all letters as unrelated in form. This highlights a key advantage of inputting pixel images into a model (similar to a retinal encoding) as opposed to abstract letter codes that lose relevant visual information.

In some respects the poorer performance of Transformers compared to CNNs is surprising given past findings that the pattern of errors observed in object recognition is more similar between visual transformers and humans compared to CNNs and humans \citep{tuli-2021}. But at the same time, our findings are consistent with the finding that CNNs provide the best predictions of neural activation in the ventral visual stream during object recognition as measured by Brain-Score \citep{schrimpf-2018} and other brain benchmarks. Indeed, visual transformers (similar to the ones we tested here) do much worse on Brain-Score compared to CNNs, with the top performing transformer model performing outside the top-100 models on the current Brain-Score leaderboard. To the extent that better performance on Brain-Score reflects a greater similarity between DNNs and humans, our finding that CNNs do a better job in accounting for masked form priming makes sense. But what specific features of CNNs lead to better performance is currently unclear.

Our findings are also consistent with recent work by \cite{hannagan-2021} who found that a CNN trained to classify images of words and objects showed a number of hallmark findings of human visual word recognition. This includes CNNs learning units that are both selective for words (analogous to neurons in the visual word form area) as well as invariant to letter size, font, or case. Furthermore, lesions to these units led to selective difficulties in identifying words (analogous to dyslexia following lesions to the visual word form area). Interestingly, the authors also provided evidence that the CNN learned a complex combination of position specific letter codes as well as bigram representations. It seems that these learned representations are also able to account for a substantial amount of of form priming effects observed in humans. The observation that CNNs not only account for a range of empirical phenomena regarding human visual word identification but, in addition, perform well on various brain benchmarks for visual object identification lends some support to the ``recycling hypothesis'' according to which a subpart of the ventral visual pathway initially involved in face and object recognition is repurposed for letter recognition \citep{hannagan-2021}.

Despite these successes, it is important to note that there is a growing number of studies highlighting that CNNs fail to capture most key low-level, mid-vision, and high-level vision findings reported in psychology \citep{bowers-inpress}.  Indeed, most CNNs that perform well on brain-score do not even classify objects on the basis of shape, and rather, classify objects on the basis of texture \citep{geirhos-2018}. And when models are trained to have a shape bias when classifying objects they have a non-human shape bias \citep{malhotra-no-date}. In addition, when tested on stimuli designed to elicit Gestalt effects, most CNNs exhibit a limited ability to organise elements of a scene into a group or whole, and the grouping only occurs at the output layer, suggesting that these models may learn fundamentally different perceptual properties than humans \citep{gestalt-2022}.

How is it possible to obtain such high performance benchmarks as Brain-Score and account for a few findings from psychology? \cite{bowers-inpress} (also see \citealt{dujmovic-2021}) argued that the good performance on these benchmarks may provide a misleading estimate of CNN-human similarity with good performance reflecting two very different systems picking up on different sources of information that are correlated with each other. For instance, it is possible that texture representations in CNNs are used not only to identify objects but also to predict the neural activation of the ventral visual system that identifies objects based on shape. Indeed, \citealt{dujmovic-2021} have run simulations showing that CNNs that are designed to recognise objects in a very different way nevertheless can support good predictions of brain activations based on confounds that are commonplace in image datasets.

In some ways, this makes the current findings all the more impressive, as the CNNs are doing reasonably well at accounting for a complex set of priming conditions that were specifically designed to contrast different hypotheses regarding orthographic coding schemes. Of course, there were some notable failures in the CNNs' accounting for form priming (most notably, all the CNNs underestimated the amount of priming in the half condition in which the primes were composed of the first three or final three letters of the target), and there may be many additional priming conditions that prove problematic for CNNs. Nevertheless, the current results do highlight that CNNs should be given more attention in psychology as models of human visual word recognition. It is possible that developing new CNN architectures motivated by biological and psychological findings and adopting more realistic training conditions will lead to even more impressive performance and new insights into human visual word identification.

\section{Conflict of Interest Statement}

The authors declare that the research was conducted in the absence of any commercial or financial relationships that could be construed as a potential conflict of interest.

\section{Declarations}

\subsection{Author Contributions}
J.S.B and V.B were responsible for the design and supervision of the study.  D.Y specified the statistical approach and wrote the scripts for data generation and analyses. All authors contributed to the analysis of the data and interpretation and writing of the paper.

\subsection{Data Availability}
The code that generate the data that support the findings of this study are available at: https://github.com/Don-Yin/Orthographic-DNN

\subsection{Code availability}
The code that support the findings of this study are available at: \\ https://github.com/Don-Yin/Orthographic-DNN

\subsection{Ethics statement}
The present study was approved by the School of Psychological ScienceResearch Ethics Committee.

\subsection{Consent to Publish}
We certify that the paper contains no personal information (names, initials, or any other information which could identify an individual person) that would infringe upon that person's right to privacy.

\subsection{Funding}
This project has received funding from the European Research Council (ERC) under the European Union’s Horizon 2020 research and innovation programme (grant agreement No 741134).

\subsection{Consent to Participate}
Not Applicable.

\clearpage
\bibliographystyle{plainnat}
\bibliography{references}

\newpage
\appendix

\begin{landscape}
    \section{\\Similarity Values}\label{appendix: data}
    \begin{table}[!ht]
        \centering
        \resizebox{1.4\textwidth}{!}{%
            \begin{tabular}{lllllllllllll}
                \hline
                \multicolumn{1}{c|}{\multirow{2}{*}{Prime Type}}  &
                \multicolumn{1}{c|}{\multirow{2}{*}{Priming-ARB}} &
                \multicolumn{7}{c|}{CNNs}                         &
                \multicolumn{4}{c}{ViTs}                                                                                                               \\
                \multicolumn{1}{c|}{}                             &
                \multicolumn{1}{c|}{}                             &
                \multicolumn{1}{c}{AlexNet}                       &
                \multicolumn{1}{c}{DenseNet169}                   &
                \multicolumn{1}{c}{Efficientnet-B1}               &
                \multicolumn{1}{c}{ResNet50}                      &
                \multicolumn{1}{c}{ResNet101}                     &
                \multicolumn{1}{c}{VGG16}                         &
                \multicolumn{1}{c|}{VGG19}                        &
                \multicolumn{1}{c}{ViT-B/16}                      &
                \multicolumn{1}{c}{ViT-B/32}                      &
                \multicolumn{1}{c}{ViT-L/16}                      &
                \multicolumn{1}{c}{ViT-L/32}                                                                                                           \\ \hline
                ID                                                & 42.69 & 1.00 & 1.00 & 1.00 & 1.00 & 1.00 & 1.00 & 1.00 & 1.00 & 1.00 & 1.00 & 1.00 \\
                DL-1F                                             & 34.23 & 0.83 & 0.72 & 0.66 & 0.89 & 0.87 & 0.78 & 0.78 & 0.96 & 0.90 & 0.81 & 0.91 \\
                IL-1F                                             & 33.66 & 0.86 & 0.75 & 0.72 & 0.91 & 0.88 & 0.80 & 0.82 & 0.96 & 0.89 & 0.79 & 0.90 \\
                TL56                                              & 32.46 & 0.88 & 0.80 & 0.82 & 0.92 & 0.91 & 0.83 & 0.84 & 0.99 & 0.97 & 0.94 & 0.96 \\
                TL-M                                              & 31.42 & 0.95 & 0.81 & 0.86 & 0.92 & 0.90 & 0.89 & 0.90 & 0.99 & 0.97 & 0.94 & 0.97 \\
                DL-1M                                             & 29.56 & 0.90 & 0.73 & 0.70 & 0.88 & 0.86 & 0.83 & 0.83 & 0.96 & 0.91 & 0.82 & 0.91 \\
                SN-F                                              & 29.45 & 0.87 & 0.75 & 0.73 & 0.92 & 0.89 & 0.77 & 0.76 & 0.98 & 0.97 & 0.91 & 0.96 \\
                SN-I                                              & 29.16 & 0.87 & 0.68 & 0.60 & 0.87 & 0.83 & 0.69 & 0.70 & 0.97 & 0.96 & 0.89 & 0.95 \\
                TL12                                              & 29.03 & 0.91 & 0.75 & 0.74 & 0.89 & 0.86 & 0.79 & 0.81 & 0.98 & 0.97 & 0.93 & 0.96 \\
                IL-1M                                             & 29.00 & 0.91 & 0.78 & 0.73 & 0.90 & 0.88 & 0.88 & 0.87 & 0.96 & 0.89 & 0.80 & 0.91 \\
                IL-1I                                             & 26.67 & 0.87 & 0.70 & 0.62 & 0.88 & 0.84 & 0.77 & 0.78 & 0.95 & 0.89 & 0.79 & 0.91 \\
                SUB3                                              & 25.83 & 0.51 & 0.37 & 0.21 & 0.67 & 0.63 & 0.44 & 0.45 & 0.92 & 0.89 & 0.77 & 0.84 \\
                IL-2MR                                            & 25.48 & 0.81 & 0.62 & 0.50 & 0.83 & 0.79 & 0.74 & 0.72 & 0.93 & 0.88 & 0.75 & 0.85 \\
                DL-2M                                             & 24.91 & 0.76 & 0.54 & 0.42 & 0.77 & 0.73 & 0.68 & 0.67 & 0.94 & 0.90 & 0.79 & 0.87 \\
                SN-M                                              & 22.68 & 0.91 & 0.73 & 0.70 & 0.88 & 0.86 & 0.82 & 0.80 & 0.98 & 0.96 & 0.90 & 0.96 \\
                N1R                                               & 21.77 & 0.88 & 0.68 & 0.65 & 0.85 & 0.82 & 0.78 & 0.77 & 0.97 & 0.95 & 0.88 & 0.95 \\
                NATL-24/35                                        & 20.20 & 0.88 & 0.66 & 0.73 & 0.84 & 0.81 & 0.78 & 0.78 & 0.98 & 0.96 & 0.92 & 0.96 \\
                IL-2M                                             & 19.42 & 0.82 & 0.64 & 0.51 & 0.84 & 0.79 & 0.77 & 0.75 & 0.93 & 0.88 & 0.75 & 0.85 \\
                T-All                                             & 16.77 & 0.75 & 0.52 & 0.52 & 0.76 & 0.72 & 0.59 & 0.60 & 0.95 & 0.91 & 0.81 & 0.90 \\
                DSN-M                                             & 14.94 & 0.84 & 0.60 & 0.52 & 0.80 & 0.77 & 0.72 & 0.69 & 0.97 & 0.94 & 0.85 & 0.95 \\
                RH                                                & 13.44 & 0.68 & 0.43 & 0.38 & 0.72 & 0.67 & 0.52 & 0.51 & 0.96 & 0.92 & 0.83 & 0.91 \\
                NATL25                                            & 9.91  & 0.83 & 0.57 & 0.65 & 0.79 & 0.78 & 0.70 & 0.69 & 0.97 & 0.96 & 0.90 & 0.95 \\
                IH                                                & 8.90  & 0.61 & 0.36 & 0.33 & 0.69 & 0.64 & 0.47 & 0.46 & 0.94 & 0.90 & 0.79 & 0.88 \\
                TH                                                & 8.80  & 0.51 & 0.32 & 0.29 & 0.68 & 0.63 & 0.43 & 0.40 & 0.93 & 0.90 & 0.77 & 0.87 \\
                ALD-PW                                            & 4.80  & 0.52 & 0.27 & 0.09 & 0.60 & 0.58 & 0.35 & 0.31 & 0.92 & 0.88 & 0.74 & 0.86 \\
                RF                                                & 2.86  & 0.69 & 0.45 & 0.49 & 0.74 & 0.71 & 0.59 & 0.57 & 0.96 & 0.93 & 0.84 & 0.91 \\
                EL                                                & 2.34  & 0.71 & 0.42 & 0.29 & 0.69 & 0.67 & 0.54 & 0.50 & 0.95 & 0.91 & 0.79 & 0.91 \\
                ALD-ARB                                           & 0.00  & 0.53 & 0.28 & 0.12 & 0.62 & 0.59 & 0.36 & 0.34 & 0.93 & 0.89 & 0.74 & 0.86 \\ \hline
            \end{tabular}%
        }
        \caption{Priming-ARB: index of priming size (ms); CNNs, ViTs: mean cosine similarity.}
        \label{tab:similarity 1}
    \end{table}
\end{landscape}

\begin{table}[!ht]
    \centering
    \resizebox{\textwidth}{!}{%
        \begin{tabular}{lllllllllll}
            \hline
            \multicolumn{1}{c|}{\multirow{2}{*}{Prime Type}} &
            \multicolumn{5}{c|}{Coding Schemes}              &
            \multicolumn{3}{c|}{Priming Models}              &
            \multicolumn{1}{c}{Baseline}                                                                                           \\
            \multicolumn{1}{c|}{}                            &
            \multicolumn{1}{c}{Absolute}                     &
            \multicolumn{1}{c}{SC}                           &
            \multicolumn{1}{c}{BOB}                          &
            \multicolumn{1}{c}{OOB}                          &
            \multicolumn{1}{c|}{SOB}                         &
            \multicolumn{1}{c}{SCM}                          &
            \multicolumn{1}{c}{IA}                           &
            \multicolumn{1}{c|}{LTRS}                        &
            \multicolumn{1}{c}{PixCS}                                                                                              \\ \hline
            ID                                               & 1.00 & 1.00 & 1.00 & 1.00 & 1.00 & -63.00  & -119.00 & 53.00 & 1.00 \\
            DL-1F                                            & 0.83 & 0.75 & 0.75 & 0.79 & 0.71 & -82.00  & -117.00 & 30.00 & 0.14 \\
            IL-1F                                            & 1.00 & 0.88 & 1.00 & 1.00 & 1.00 & -76.00  & -121.00 & 21.00 & 0.14 \\
            TL56                                             & 0.67 & 0.81 & 0.83 & 0.83 & 0.94 & -87.00  & -133.00 & 23.00 & 0.82 \\
            TL-M                                             & 0.67 & 0.93 & 0.86 & 0.79 & 0.93 & -75.00  & -133.00 & 26.67 & 0.82 \\
            DL-1M                                            & 0.42 & 0.83 & 0.63 & 0.66 & 0.71 & -82.00  & -130.00 & 40.25 & 0.14 \\
            SN-F                                             & 0.83 & 0.75 & 0.75 & 0.79 & 0.71 & -95.00  & -124.00 & 25.00 & 0.62 \\
            SN-I                                             & 0.83 & 0.75 & 0.75 & 0.79 & 0.30 & -98.00  & -124.00 & 20.00 & 0.59 \\
            TL12                                             & 0.67 & 0.81 & 0.83 & 0.83 & 0.66 & -88.00  & -133.00 & 19.00 & 0.82 \\
            IL-1M                                            & 0.50 & 0.90 & 0.75 & 0.85 & 0.96 & -78.00  & -139.00 & 25.00 & 0.14 \\
            IL-1I                                            & 0.00 & 0.88 & 1.00 & 1.00 & 0.61 & -76.00  & -143.00 & 17.00 & 0.14 \\
            SUB3                                             & 0.25 & 0.50 & 0.25 & 0.36 & 0.29 & -105.00 & -130.00 & 16.50 & 0.16 \\
            IL-2MR                                           & 0.50 & 0.73 & 0.58 & 0.75 & 0.96 & -93.00  & -140.00 & 16.00 & 0.32 \\
            DL-2M                                            & 0.33 & 0.57 & 0.25 & 0.33 & 0.57 & -106.00 & -125.00 & 36.00 & 0.31 \\
            SN-M                                             & 0.83 & 0.88 & 0.63 & 0.61 & 0.75 & -84.00  & -124.00 & 25.00 & 0.58 \\
            N1R                                              & 0.67 & 0.84 & 0.58 & 0.54 & 0.76 & -89.00  & -133.00 & 20.00 & 0.53 \\
            NATL-24/35                                       & 0.67 & 0.82 & 0.67 & 0.44 & 0.86 & -86.00  & -133.00 & 20.50 & 0.74 \\
            IL-2M                                            & 0.50 & 0.73 & 0.58 & 0.75 & 0.96 & -93.00  & -140.00 & 16.00 & 0.31 \\
            T-All                                            & 0.00 & 0.48 & 0.67 & 0.44 & 0.59 & -106.00 & -143.00 & 10.00 & 0.47 \\
            DSN-M                                            & 0.67 & 0.75 & 0.25 & 0.31 & 0.62 & -103.00 & -133.00 & 13.00 & 0.42 \\
            RH                                               & 0.33 & 0.39 & 0.25 & 0.18 & 0.38 & -105.00 & -141.00 & 9.00  & 0.48 \\
            NATL25                                           & 0.67 & 0.76 & 0.42 & 0.32 & 0.78 & -103.00 & -133.00 & 18.00 & 0.65 \\
            IH                                               & 0.00 & 0.31 & 0.33 & 0.38 & 0.46 & -107.00 & -142.00 & 9.00  & 0.38 \\
            TH                                               & 0.00 & 0.38 & 0.50 & 0.73 & 0.33 & -107.00 & -143.00 & 8.00  & 0.35 \\
            ALD-PW                                           & 0.00 & 0.00 & 0.00 & 0.00 & 0.00 & -109.00 & -144.00 & 0.00  & 0.23 \\
            RF                                               & 0.33 & 0.45 & 0.08 & 0.17 & 0.70 & -106.00 & -141.00 & 10.00 & 0.48 \\
            EL                                               & 0.33 & 0.50 & 0.00 & 0.00 & 0.21 & -107.00 & -140.00 & 3.00  & 0.31 \\
            ALD-ARB                                          & 0.00 & 0.00 & 0.00 & 0.00 & 0.00 & -108.00 & -142.00 & 0.00  & 0.26 \\ \hline
        \end{tabular}%
    }
    \caption{Coding schemes, LTRS: predicted match value; SCM and IA: predicted mean reaction time (ms). In order to obtain a positive correlation value, the SCM and IA values are made negative.}
    \label{tab:similarity 2}
\end{table}

\section{\\Data Generation}\label{appendix: data generation}
\begin{figure}[!ht]
    \centering
    \adjustbox{scale=1.3,center}{
        \begin{tikzcd}
            \text{abduct} \arrow[d, "\text{Set Font}"']                                     & \text{abound} \arrow[d, "\text{Set Font}"']                                     & \text{abrupt} \arrow[d, "\text{Set Font}"']                                     & ... \\
            {\includegraphics[scale=0.3]{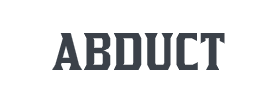}} \arrow[d, "\text{Draw}"']         & {\includegraphics[scale=0.3]{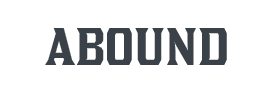}} \arrow[d, "\text{Draw}"']         & {\includegraphics[scale=0.3]{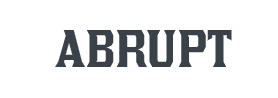}} \arrow[d, "\text{Draw}"']         & ... \\
            {\includegraphics[scale=0.25]{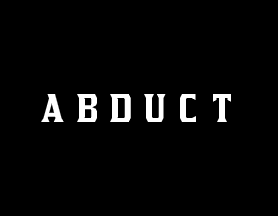}} \arrow[d, "\text{Rotate}"']      & {\includegraphics[scale=0.25]{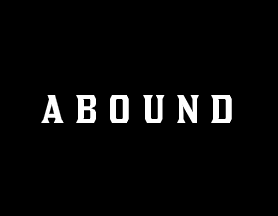}} \arrow[d, "\text{Rotate}"']      & {\includegraphics[scale=0.25]{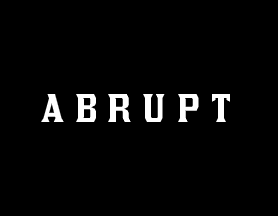}} \arrow[d, "\text{Rotate}"']      & ... \\
            {\includegraphics[scale=0.25]{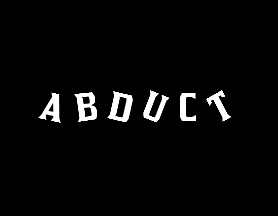}} \arrow[d, "\text{Translate}"'] & {\includegraphics[scale=0.25]{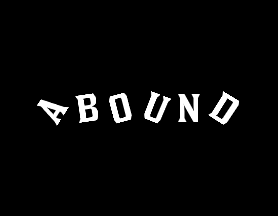}} \arrow[d, "\text{Translate}"'] & {\includegraphics[scale=0.25]{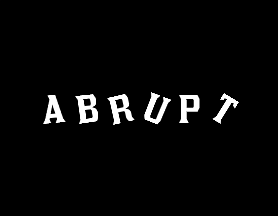}} \arrow[d, "\text{Translate}"'] & ... \\
            {\includegraphics[scale=0.25]{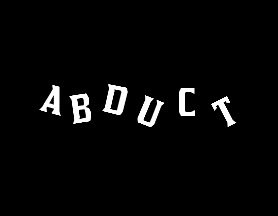}}                             & {\includegraphics[scale=0.25]{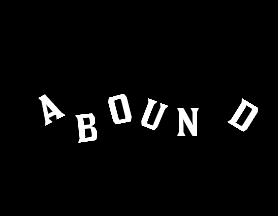}}                             & {\includegraphics[scale=0.25]{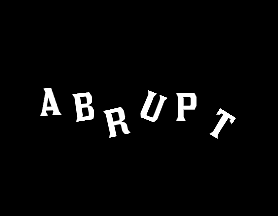}}                             & ...
        \end{tikzcd}}
    \caption{Illustration of the process of generating training/validation images from target words (e.g., ‘ABDUCT’). The word is drawn with a random font and size. Each letter is then rotated and translated randomly. This procedure is applied on each target word to generate 6,000 images, resulting in 6,000,000 images (1,000 words $\times$ 6,000 images).}
    \label{dig:process generation}
\end{figure}

Figure \ref{dig:process generation} illustrates how the 1,000 target words are used to generate 6,000 images per word algorithmically, which involves the following steps:

\begin{enumerate}
    \item Apply one of the ten common fonts (e.g., Arial). The complete list can be found at the Github repository of the current study \footnote{https://github.com/Don-Yin/Orthographic-DNN}.
    \item Apply one of ten sizes selected from $
              \{x \in 2 Z: 18 \leq x<38\} $
    \item Draw the target word as an image.
    \item Add random rotation to individual letters using an angle determined by the normal distribution $N\left(0, \frac{2 \pi}{45}\right)$.
    \item Add random translation to individual letters so that the updated letter coordinates $(x, y)$ meet the condition shown in Equation \ref{eq:translation}, where $(a, b)$ are the letter’s initial coordinates.
    \item Transform the image into grayscale.
    \item Resize image to 224 $\times$ 224 pixels.
\end{enumerate}

\begin{equation} \label{eq:translation}
    \begin{aligned}
        \left(0.8 \times \sqrt{\text { height }_{\text{letter }}^{2}+width_{\text {letter}}^{2}}\right)^{2} \geq(x-b)^{2}+(y-a)^{2} \quad \\
        \forall (a, b) \in \text{Bounding Circle}
    \end{aligned}
\end{equation}

These operations ensure sufficient variation between the generated images to simulate human perceptual invariance to rotation and spacing. At the end of this process, 6,000,000 images are generated (1,000 words $\times$ 6,000 images). 5,000,000 images are used for training and the remaining 1,000,000 are used for validating the DNN models’ performance.

\begin{landscape}
    \begin{figure}[!hp]
        \centering
        \includegraphics[width=0.80\linewidth,trim={9cm 0 9cm 6cm},clip]{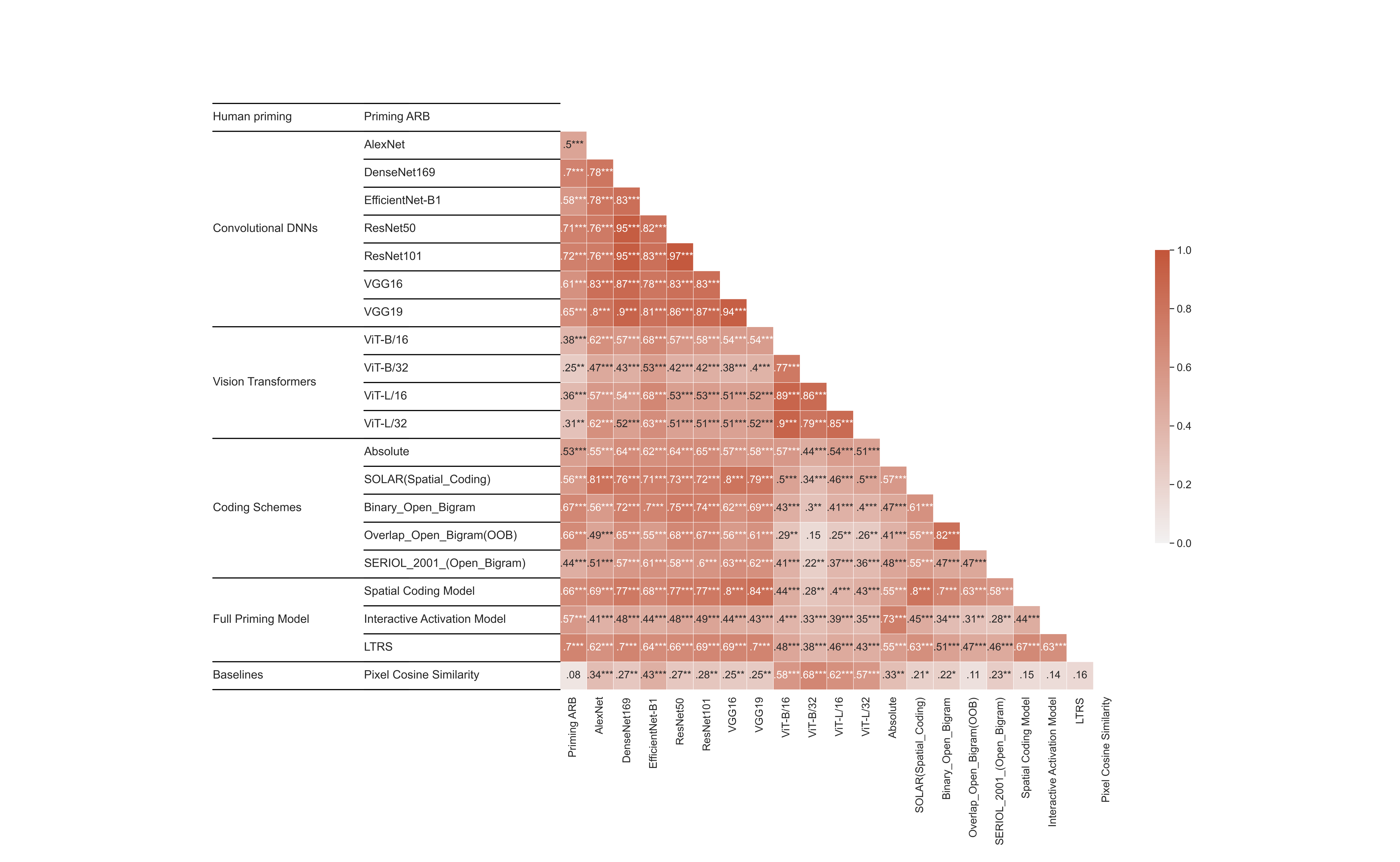}
        \caption{Pairwise Correlation Matrix Between Human Priming Data, the DNN Models and Other Similarity Metrics. The values represent Kendall’s $\tau$ correlation coefficients. Priming-ARB is the human priming size using the arbitrary unrelated prime condition as the baseline; pixCS is the Pixel cosine similarity; SCM is the Spatial Coding Model. It should be noted that, in order to obtain a positive correlation value, the LDist and SCM's values are made negative.\\
            ${ }^{*} p<.05 .{ }^{* *} p<.01 .{ }^{* * *} p<.001$}
        \label{fig:correlation main}
    \end{figure}
\end{landscape}

\begin{table}[!ht]
    \centering
    \resizebox{\textwidth}{!}{%
        \begin{tabular}{lllllllll}
            \hline
            \multirow{2}{*}{Architecture}          &
            \multirow{2}{*}{Family}                &
            \multirow{2}{*}{Model}                 &
            \multicolumn{4}{l}{Neural Benchmarks}  &
            \multirow{2}{*}{Behavioural Benchmark} &
            \multirow{2}{*}{Brain-Score}                                                                                                 \\
                                                   &                         &                 & V1   & V2   & V4   & IT   &      &      \\ \hline
            \multirow{7}{*}{Convolutional}         & AlexNet                 & AlexNet         & 0.51 & 0.35 & 0.44 & 0.36 & 0.37 & 0.41 \\
                                                   & DenseNet                & Densenet169     & 0.49 & 0.32 & 0.5  & 0.38 & 0.54 & 0.45 \\
                                                   & EfficientNet            & EfficientNet-B1 & 0.49 & 0.33 & 0.49 & 0.38 & 0.55 & 0.45 \\
                                                   & \multirow{2}{*}{ResNet} & ResNet50        & 0.51 & 0.32 & 0.49 & 0.41 & 0.53 & 0.45 \\
                                                   &                         & ResNet101       & 0.49 & 0.34 & 0.49 & 0.4  & 0.56 & 0.46 \\
                                                   & \multirow{2}{*}{VGG}    & VGG16           & 0.43 & 0.51 & 0.34 & 0.48 & 0.37 & 0.43 \\
                                                   &                         & VGG19           & 0.54 & 0.34 & 0.47 & 0.36 & 0.49 & 0.44 \\ \cline{1-2}
            \multirow{4}{*}{Transformer}           & \multirow{4}{*}{ViT}    & ViT-B/16        & 0.12 & 0.26 & 0.12 & 0.08 & 0.34 & 0.18 \\
                                                   &                         & ViT-B/32        & 0.14 & 0.29 & 0.13 & 0.09 & 0.28 & 0.18 \\
                                                   &                         & ViT-L/16        & 0.11 & 0.27 & 0.12 & 0.10 & 0.33 & 0.19 \\
                                                   &                         & ViT-L/32        & 0.15 & 0.30 & 0.13 & 0.09 & 0.29 & 0.19 \\ \hline
        \end{tabular}%
    }
    \caption{The DNNs models and their Brain-Scores (Schrimpf et al., 2018). Each neural benchmark corresponds to a specific area of the visual system. The mean value of benchmarks is calculated to represent an overall `Brain-Score'.}
    \label{tab:model info}
\end{table}
Figure \ref{fig:correlation main} illustrates the pair-wise Kendall’s correlation matrix between the human priming-ARB and the DNN models and baseline similarity metrics (similarity values are listed in \ref{appendix: data}).

\end{document}